\documentclass[letterpaper, 10 pt, conference]{ieeeconf}  % Comment this line out if you need a4paper

\IEEEoverridecommandlockouts                              % This command is only needed if 
\usepackage{times}
\usepackage{amssymb}
\usepackage{amsmath}
\usepackage{amsthm}
\usepackage{graphics}
\usepackage{epsfig}
\usepackage{multirow}
\usepackage{nicefrac}
\usepackage{cases}          % For multi-line equations with cases and numbers
\usepackage{subcaption}
\usepackage{mathtools}
\usepackage{xcolor}
\usepackage{algorithm}
\usepackage{algorithmicx}
\usepackage{algpseudocode}
\usepackage{balance}
\usepackage{gensymb}
\usepackage{comment}
\usepackage{hyperref}
\usepackage[capitalise]{cleveref}
\usepackage{bm}

\DeclareMathAlphabet{\pazocal}{OMS}{zplm}{m}{n}
\newcommand{\mat}[0]{\begin{bmatrix}}
\newcommand{\mate}[0]{\end{bmatrix}}
\newcommand{\va}{\mathbf{a}}

\newcommand{\vg}{\mathbf{g}}
\newcommand{\vh}{\mathbf{h}}

\newcommand{\vp}{\mathbf{p}}

\newcommand{\vr}{\mathbf{r}}
\newcommand{\vs}{\mathbf{s}}
\newcommand{\vS}{\mathbf{S}}

\newcommand{\vu}{\mathbf{u}}

\newcommand{\vx}{\mathbf{x}}
\newcommand{\vX}{\mathbf{X}}

\newcommand{\cB}{\mathcal{B}}

\newcommand{\cD}{\mathcal{D}}

\newcommand{\cN}{\mathcal{N}}

\newcommand{\cS}{\mathcal{S}}

\newcommand{\cU}{\mathcal{U}}

\newcommand{\cX}{\mathcal{X}}

\newcommand{\R}{\mathbb{R}}

\newcommand{\hp}{\hat{\mathbf{p}}}
\newcommand{\hv}{\hat{\mathbf{v}}}

\newcommand\norm[1]{\left\|#1\right\|}              % Big norm ||#||
\DeclareMathOperator*{\argmin}{arg\,min}            % argmin
\newcommand{\la}{\leftarrow}

\newtheorem{definition}{\textbf{\textit{Definition}}}

\usepackage{xcolor}

\makeatletter
\let\NAT@parse\undefined
\makeatother
\usepackage{hyperref}
\usepackage[sort, compress]{cite}
\hypersetup{linkbordercolor=green}
\usepackage{array}
\newcolumntype{P}[1]{>{\centering\arraybackslash}p{#1}}
\title{\LARGE \bf
Where to go next: Learning a Subgoal Recommendation Policy \\ for Navigation in Dynamic Environments
}
 
\author{Bruno Brito$^1$, Michael Everett$^2$, Jonathan P. How$^2$ and Javier Alonso-Mora$^1$ % <-this % stops a space
\thanks{This work was supported by the European Union’s Horizon 2020 research and innovation 
programme under grant agreement No. 101017008, the Amsterdam Institute for Advanced Metropolitan Solutions, the Netherlands Organisation for Scientific Research (NWO) domain Applied Sciences (Veni 15916), and Ford Motor Company.}% <-this % stops a space
\thanks{$^1$The authors are with the Cognitive Robotics (CoR) department,
        Delft University of Technology, 2628 CD Delft, The Netherlands
    {\tt\small \{bruno.debrito, j.alonsomora\}@tudelft.nl}}%
\thanks{$^2$The authors are with Massachusetts Institute of Technology, Aerospace
Controls Laboratory, Cambridge, MA, USA. {\tt\small \{mfe, jhow\}@mit.edu}}
\thanks{\textbf{Code: }{\url{https://github.com/tud-amr/go-mpc.git}}}
\thanks{\textbf{Video: }{\url{https://youtu.be/sZBbWMnwle8}}}
}

\begin{document}

%%%%%%%%%%%%%%%%%%%%%%%%%%%%%%%%%%%%%%%%%%%%%%%%%%%%%%%%%%%%%%%%%%%%%%%%%%%%%%%%

\maketitle
\thispagestyle{empty}
\pagestyle{empty}
%%%%%%%%%%%%%%%%%%%%%%%%%%%%%%%%%%%%%%%%%%%%%%%%%%%%%%%%%%%%%%%%%%%%%%%%%%%%%%%%

\begin{abstract}

% This paper proposes a new framework for safe robot navigation among pedestrians, by introducing 

Robotic navigation in environments shared with other robots or humans remains challenging because the intentions of the surrounding agents are not directly observable and the environment conditions are continuously changing. 
Local trajectory optimization methods, such as model predictive control (MPC), can deal with those changes but require global guidance, which is not trivial to obtain in crowded scenarios.
This paper proposes to learn, via deep Reinforcement Learning (RL), an interaction-aware policy that provides long-term guidance to the local planner. In particular, in simulations with cooperative and non-cooperative agents, we train a deep network to recommend a subgoal for the MPC planner. The recommended subgoal is expected to help the robot in making progress towards its goal and accounts for the expected interaction with other agents. Based on the recommended subgoal, the MPC planner then optimizes the inputs for the robot satisfying its kinodynamic and collision avoidance constraints.
Our approach is shown to substantially improve the navigation performance in terms of
number of collisions as compared to prior MPC frameworks, and in terms of both travel time and number of collisions compared to deep RL methods in cooperative, competitive and mixed multiagent scenarios.

\end{abstract}

%!TEX root=../main.tex

% The new proposed method links re-planning techniques on global scale with an optimization based controller approach on local scale to generate safe and smooth motion. Having both the ability of optimization-based planners that execute appropriate reactive behavior by explicitly incorporating constraints, as well as the efficiency and global convergence of search-based re-planning techniques, advantages of both methods are combined. 
%%%%%%%%%%%%%%%%%%%%%%%%%%%%%%%%%%%%%%%%%%%%%%%%%%%%%%%%%%%%%%%%%%%%%%%%%%%%%%%%
\section{INTRODUCTION}

Autonomous robot navigation in crowds remains difficult due to the interaction effects among navigating agents.
Unlike multi-robot environments, robots operating among pedestrians require decentralized algorithms that can handle a mixture of other agents' behaviors without depending on explicit communication between agents.
% While pedestrians can safely navigate through these environments without explicit communication, decentralized and non-communicating methods for motion planning algorithms is necessary.

Several state-of-the-art collision avoidance methods employ model-predictive control (MPC) with online optimization to compute motion plans that are guaranteed to respect important constraints \cite{paden2016survey}.
These constraints could include the robot's nonlinear kino-dynamics model or collision avoidance of static obstacles and other dynamic, decision-making agents (e.g., pedestrians).
Although online optimization becomes less computationally practical for extremely dense scenarios, modern solvers enable real-time motion planning in many situations of interest \cite{brito2019model}.

A key challenge is that the robot's global goal is often located far beyond the planning horizon, meaning that a local subgoal or cost-to-go heuristic must be specified instead.
This is straightforward in a static environment (e.g., using euclidean/diffusion~\cite{chen2016motion} distance), but the presence interactive agents makes it difficult to quantify which subgoals will lead to the global goal quickest.
A body of work addresses this challenge with deep reinforcement learning (RL), in which agents learn a model of the long-term cost of actions in an offline training phase (usually in simulation) \cite{chen2017decentralized,chen2019crowd,everett2019collision,fan2020distributed}.
The learned model is fast-to-query during online execution, but the way learned costs/policies have been used to date does not provide guarantees on collision avoidance or feasibility with respect to the robot dynamics.

\begin{figure}
    \centering
    \includegraphics[width=0.8\columnwidth]{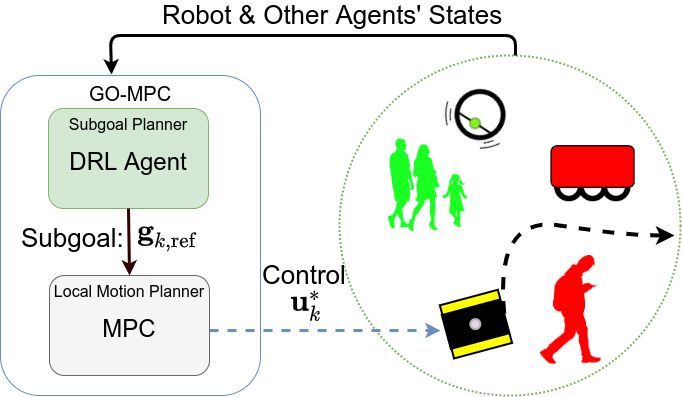}
    \caption{Proposed navigation architecture. The subgoal planner observes the environment and suggests the next subgoal position to the local motion planner, the MPC. The MPC then computes a local trajectory and the robot executes the next optimal control command, which minimizes the distance to the provided position reference while respecting collision and kinodynamic constraints.}
    \label{fig:architecture_proposed}
    \vspace{-0.2in}
\end{figure}

%In this paper, we introduce a Goal Oriented Model Predictive Control (GO-MPC) for navigation among cooperative and non-cooperative agents, as depicted in Fig~\ref{fig:architecture_proposed}. We employ RL to learn a policy providing an intermediate goal position to a MPC. Then, the MPC computes driving the 

In this paper, we introduce Goal Oriented Model Predictive Control (GO-MPC), which enhances state-of-art online optimization-based planners with a learned global guidance policy.
%This work enhances state-of-art online optimization-based planners with a learned global guidance policy.
In an offline RL training phase, an agent learns a policy that uses the current world configuration (the states of the robot and other agents, and a global goal) to recommend a local subgoal for the MPC, as depicted in~\cref{fig:architecture_proposed}.
Then, the MPC generates control commands ensuring that the robot and collision avoidance constraints are satisfied (if a feasible solution is found) while making progress towards the suggested subgoal.
Our approach maintains the kino-dynamic feasibility and collision avoidance guarantees inherent in an MPC formulation, while improving the average time-to-goal and success rate by leveraging past experience in crowded situations.

The main contributions of this work are:
\begin{itemize}
    %\item A trajectory optimization framework for motion planning in crowded environments that utilizes a learned cost for guidance and provides dynamic feasibility and collision avoidance guarantees;
    \item A goal-oriented Model Predictive Control method (GO-MPC) for navigation among interacting agents, which utilizes a learned global guidance policy (recommended subgoal) in the cost function and ensures that dynamic feasibility and collision avoidance constraints are satisfied when a feasible solution to the optimization problem is found;
    %\item An algorithm to train an RL agent whose actions are cascaded with an optimization-based controller.
    \item An algorithm to train an RL agent jointly with an optimization-based controller in mixed environments, which is directly applicable to real-hardware, reducing the sim to real gap.
    %\item leveraging the ability to learn from past experiences to improve navigation the cost-to-go or terminal constraint.
\end{itemize}

Finally, we present simulation results demonstrating an improvement over several state-of-art methods in challenging scenarios with realistic robot dynamics and a mixture of cooperative and non-cooperative neighboring agents. Our approach shows different navigation behaviors: navigating through the crowd when interacting with cooperative agents, avoiding congestion areas when non-cooperative agents are present and enabling communication-free decentralized multi-robot collision avoidance.

\subsection{Related Work}

\subsubsection{Navigation Among Crowds}

%A standard architecture in crowd navigation is shown in~\cref{fig:standard_architecture}: at each timestep, sensor data and a global goal position are used to compute the robot's control sequence.
%This is often done by extracting a map of the static environment and the vehicle and pedestrian states from a perception system, then feeding those estimates into various world models to form the planning costs and constraints.
Past work on navigation in cluttered environments
often focuses on \textit{interaction models} using geometry~\cite{van2011reciprocal,van2011reciprocalacc}, physics~\cite{helbing1995social}, topologies~\cite{mavrogiannis2019multi,mavrogiannis2018social}, handcrafted functions \cite{trautman2015robot}, and cost functions \cite{kim2016socially,kim2016socially} or joint probability distributions \cite{vemula2017modeling} learned from data.
While accurate interaction models are critical for collision avoidance, this work emphasizes that the robot's performance (time-to-goal) is highly dependent on the quality of its \textit{cost-to-go model} (i.e., the module that recommends a subgoal for the local planner).
%to learn a joint distribution 
%\cite{trautman2015robot} Robot navigation in dense human crowds
%to learn a cost function from expert demonstrations
%\cite{kim2016socially} Socially adaptive path planning in human environments using inverse reinforcement learning
%\cite{} Social momentum:
%\cite{kretzschmar2016socially} Socially compliant mobile robot navigation
%\cite{vemula2017modeling} Modeling cooperative navigation in dense human crowds
Designing a useful cost-to-go model in this problem remains challenging, as it requires quantifying how ``good'' a robot's configuration is with respect to dynamic, decision-making agents.
% can we bring multi-policy \cite{galceran2017multipolicy} into this discussion? i recall it wasn't great at cost-to-go estimation.
In \cite{chen2017decentralized}, deep RL was introduced as a way of modeling cost-to-go through an offline training phase; the online execution used simple vehicle and interaction models for collision-checking. Subsequent works incorporated other interactions to generate more socially compliant behavior within the same framework~\cite{chen2017socially,chen2019crowd}.
To relax the need for simple online models, \cite{everett2019collision} moved the collision-checking to the offline training phase.
While these approaches use pre-processed information typically available from perception pipelines (e.g., pedestrian detection, tracking systems), other works proposed to learn end-to-end policies~\cite{fan2020distributed,tai2018socially}.
Although all of these RL-based approaches learn to estimate the cost-to-go, the online implementations do not provide guarantees that the recommended actions will satisfy realistic vehicle dynamics or collision avoidance constraints.
Thus, this work builds on the promising idea of learning a cost-to-go model, but we start from an inherently safe MPC formulation.

% interaction
% - dwa, rvo, sf, cornell, georges, luders

% cost-to-go
% - chen chen chen (offline cost-to-go modeling + online MPC w/ simple dynamics/interaction)
% - everett: replace online MPC w/ more intensive offline learning
% - long/fan/gail? (end-to-end)

\subsubsection{Learning-Enhanced MPC}

% What's hard about nonlinear MPC?
% - initial guess
% - model inacurracies --> suboptimal solutions in real world
% - computation time
% - writing cost function / multi-objective weights / deciding btwn cost vs. constraint
% - feasibility

%It is well known that optimisation-based methods suffer from the curse of dimensionality and thus, not scaling for large problems environments with

Outside the context of crowd navigation, numerous recent works have proposed learning-based solutions to overcome some of the known limitations of optimization-based methods (e.g., nonlinear MPC) \cite{hewing2020learning}.
% Even outside the context of crowd navigation, optimization-based methods (e.g. nonlinear MPC) has well-known practical challenges on its own. We summarize recent learning-based solutions to overcome some of the known limitations.%MPC's pain points.
For example, solvers are often sensitive to the quality of the \textit{initial guess} hence, \cite{mansard2018using} proposes to learn a policy from data that efficiently ``warm-starts'' a MPC.
\textit{Model inaccuracies} can lead to sub-optimal MPC solution quality; \cite{bellegarda2019combining} proposes to learn a policy by choosing between two actions with the best expected reward at each timestep: one from model-free RL and one from a model-based trajectory optimizer. Alternatively, RL can be used to optimize the weights of an MPC-based Q-function approximator or to update a robust MPC parametrization \cite{zanon2020safe}.
%Alternatively, \cite{zanon2019practical} uses RL to optimize the weights of an MPC-based Q-function approximator, which ensures that estimates of the cost-to-go are based on future trajectories that respect state/input constraints.
When the model is completely unknown, \cite{nagab2017neural} shows a way of learning a dynamics model to be used in MPC.
\textit{Computation time} is another key challenge: \cite{zhong2013value} learns a cost-to-go estimator to enable shortening of the planning horizons without sacrificing much solution quality, although their approach differs from this work as it uses local and linear function approximators which limits its applicability to high-dimensional state spaces.%does not use RL.
%\begin{figure}
%    \centering
%    \includegraphics[page=2,width=\columnwidth]{images/mpc_rl_front.png}
%    \caption{Standard Crowd Navigation Architecture.}
%    \label{fig:standard_architecture}
%\end{figure}
The closest related works address \textit{cost tuning} with learning.
MPC's cost functions are replaced with a value function learned via RL offline in \cite{lowrey2018plan} (terminal cost) and \cite{farshidian2019deep} (stage cost).% \BF{or \textit{hyperparameter tunning} with learning a parameter policy to reduce conservativeness but keeping safety \cite{}}.
\cite{karnchanachari2020practical} deployed value function learning on a real robot outperforming an expert-tuned MPC.
While these ideas also use RL for a better cost-to-go model, this work focuses on the technical challenge of learning a subgoal policy required for navigation through crowds avoiding the approximation issues and extrapolation issues to unseen events.
Moreover, this work learns to set terminal constraints rather than setting a cost with a value function.

\subsubsection{Combining MPC with RL}

Recently, there is increasing interest on approaches combining the strengths of MPC and RL as suggested in \cite{ernst2008reinforcement}. For instance, optimization-based planning has been used to explore high-reward regions and distill the knowledge into a policy neural network, rather than a neural network policy to improve an optimization.  \cite{levine2013guided,levine2013variational,mordatch2014combining}.
%Guided policy search uses optimization-based planning to explore high-reward regions and distill the knowledge into a policy neural network \cite{levine2013guided}. Refs.~\cite{levine2013variational,mordatch2014combining} do the reverse of this work: they use optimization to improve the search for a neural network policy, rather than a neural network policy to improve an optimization. 

 %\BF{Missing references:  does something similar to our approach but the rl agent learns to select from a limited set of already pre-computed positions which may not be available on real scenario}
 
%5Model-based RL combining probabilitics dynamic models with MPC for planning have demonstrated to achieve comparable performance with model-free RL algorithms \cite{chua2018deep}.

Similar to our approach, \cite{hong2019modelbased} utilizes the RL policy during training to ensure exploration and employs a MPC to optimize sampled trajectories from the learned policy at test time. Moreover, policy networks have be used to generate proposals for a sampling-based MPC \cite{Wang2020Exploring}, or to select goal positions from a predefined set \cite{greatwood2019reinforcement}.

Nevertheless, to the extent of our knowledge, approaches combining the benefits of both optimization and learning-based methods were not explored in the context of crowd navigation. Moreover, the works exploring a similar idea of learning a cost-to-go model do not allow to explicitly define collision constraints and ensure safety. To overcome the previous issues, in this paper, we explore the idea of learning a cost-to-go model to directly generate subgoal positions, which lead to higher long-term rewards and too give the role of local collision avoidance and kinematics constraints to an optimization-based planner.
%\BF{@Michael, should we conclude something here? maybe state the differences with the previous methods? or just the drawbacks of the methods of this last sub-section?}
%\meXX{Yeah, we should summarize the key gaps. the conclusions of the three sections are basically: lack of learned cost-to-go with safety guarantees, some ideas about learning+MPC only exist in other contexts, and i'm not sure about the last one}

Such cost-to-go information can be formulated as learning a value function for the ego-agent state-space providing information which states are more valuable \cite{farshidian2019deep}. In contrast, we propose to learn a policy directly informing which actions lead to higher rewards allowing to directly incorporate the MPC controller in the training phase.

%\subsection{Contributions}
 
%This work enhances state-of-art online optimization-based planners with a learned global guidance policy.
%In an offline RL training phase, an agent learns a policy that uses the current world configuration (vehicle and pedestrian states, global goal) to recommend a local subgoal for the MPC, as depicted in~\cref{fig:architecture_proposed}.
%Then, the MPC generates control commands ensuring that the vehicle and collision avoidance constraints are satisfied while making progress towards the suggested subgoal.
%Our approach maintains the kino-dynamic feasibility and collision avoidance guarantees inherent in an MPC formulation, while improving the average time-to-goal by leveraging past experience in crowded situations. The main contributions of this work are:
%\begin{itemize}
%    \item A framework for learning-enhanced motion planning in crowded environments that provides dynamic feasibility \& collision avoidance guarantees;
%    \item An algorithm to train an RL agent whose actions are cascaded with an optimization-based controller; and
%\end{itemize}

%Simulation results demonstrating an improvement over the state-of-art methods in challenging scenarios with realistic robot dynamics and a mixture of cooperative and competitive neighboring agents.

%We evaluate experiments show that our policy learn to behave different depending on the

%avoid congestion areas
%or more aggressive among cooperative agents

\section{PRELIMINARIES}\label{sec:problem_formulation}

Throughout this paper, vectors are denoted in bold lowercase letters, $\mathbf{x}$, matrices in capital, $ M $, and sets in calligraphic uppercase, $\mathcal{S}$.  $\norm{\vx} $ denotes the Euclidean norm of $ \vx $ and $\norm{\vx}_Q = \vx^TQ\vx$ denotes the weighted squared norm. The variables $\{\vs,\va\}$ denote the state and action variables used in the RL formulation, and $\{\vx,\vu\}$ denote the control state and action commands used in the optimization problem.
% We use the upscript $0$ to denote the ego-agent, referring to the agent controlled by our algorithm, and $-i$ to refer to all other agents.

\subsection{Problem Formulation}\label{sec:problem}
 %Here, we consider a partially observable setting since we assume that we do not know the other agent's plans.

Consider a scenario where a robot must navigate from an initial position $\mathbf{p}_0$ to a goal position $\mathbf{g}$ on the plane $\mathbb{R}^2$, surrounded by $n$ non-communicating agents. At each time-step $t$, the robot first observes its state $\vs_t$ (defined in Sec.\ref{sec:rl}) and the set of the other agents states $\mathbf{S}_t=\bigcup_{i \in \{1,\dots,n \}} \vs^i_t $, then takes action $\va_t$, leading to the immediate reward $R(\vs_t,\va_t)$ and next state $\vs_{t+1}=h(\vs_t,\va_t)$, under the transition model $h$.

We use the superscript $i\in\{1,\dots,n\}$ to denote the $i$-th nearby agent and omit the superscript when referring to the robot. For each agent $i \in \{0,n\}$, $\mathbf{p}^i_t\in\mathbb{R}^2$ denotes its position, $\mathbf{v}^i_t\in\mathbb{R}^2$ its velocity %5and $a^i_t\in\mathbb{R}^2$ its acceleration 
at step $t$ relative to a inertial frame, and $r_i$ the agent radius. We assume that each agent's current position and velocity are observed (e.g., with on-board sensing) while other agents' motion intentions (e.g., goal positions) are unknown. Finally, $\pazocal{O}_t$ denotes the area occupied by the robot and $\pazocal{O}^i_t$ by each surrounding agent, at time-step $t$.% and the set $\mathcals{O}_t = \bigcup_{i \in \{1,\dots,n \}} \pazocal{O}^i_t$ the are occupied by the other agents. 
%In this work, and without a loss of generality, we assume a constant velocity model with Gaussian noise $\mathbf{\omega}_o(t) \sim \mathcal{N}(0, Q_o(t))$ in acceleration, that is, $\ddot {\mathbf{p}}_i(t) = \mathbf{\omega}_i(t)$, where $\mathbf{p}_i(t)$ is the position of obstacle $i$ at time $t$. Given the measured position data of each obstacle, we estimate their future positions and uncertainties with a linear Kalman filter~\cite{Welch1995}.
The goal is to learn a policy $\pi$ for the robot that minimizes time to goal while ensuring collision-free motions, defined as:
\begin{subequations}
\label{eq-problem}
\begin{align}
\pi^* = \underset{\pi}{\operatorname{argmax}} \quad & \mathbb{E}\left[\sum_{t=0}^{T}\gamma^tR(\mathbf{s}_{t}, \pi(\mathbf{s}_{t},\mathbf{S}_{t}^{}))\right]\nonumber \\
\label{eq:dynamics_path_parameter}\text{s.t.} \quad & \vx_{t+1} = f(\vx_{t},\boldsymbol{u}_{t}), \\
\label{eq:terminal_constraints}\quad & \boldsymbol{s}_{T} = \mathbf{g}, \\
%&\label{eq:collision_constraints} \norm{\vp,\vp_i} \geq r+r^i \hspace{5mm} \forall i \in \{1,\dots,n\},\\
&\label{eq:collision_constraints} \pazocal{O}_t(\vx_t) \cap \pazocal{O}^i_t  = \emptyset \\
% & \B(\boldsymbol{z}_k) \cap \pazocal{O}_{static} = \not \hspace{-0.05cm} 0\\
% & \B(\boldsymbol{z}_k) \cap \pazocal{O}_{dyn} = \not \hspace{-0.05cm} 0\\
&\label{eq:state_control_constraints} \vu_t \in \cU, \,\, \vs_{t} \in \cS, \,\, \vx_{t} \in \cX, \\
& \forall t \in [0,T], \quad \forall i \in \{1,\dots,n \}\nonumber
\end{align}
\end{subequations}
where (\ref{eq:dynamics_path_parameter}) are the transition dynamic constraints considering
the dynamic model $f$, (\ref{eq:terminal_constraints}) the terminal constraints, (\ref{eq:collision_constraints}) the collision avoidance constraints and $\cS$, $\cU$ and $\cX$ are the set of admissible states, inputs (e.g., to limit the robot's maximum speed) and the set of admissible control states, respectively. Note that we only constraint the control states of the robot. Moreover, we assume other agents have various behaviors (e.g., cooperative or non-cooperative): each agent samples a policy from a closed set $\mathcal{P} = \{\pi_1,\dots,\pi_m\}$ (defined in Sec.\ref{sec:others}) at the beginning of each episode. 

%\begin{itemize}
%    \item \meXX{Should we name the result of the argmax? $\pi^*=\argmax...$? only necessary if we want to refer to that policy later on}
%\end{itemize}

%\subsection{PPO}\label{sec:ppo}
% On-policy reinforcement learning methods enable simpler integration with other experts (e.g. trajectory optimization methods).

%\meXX{Do we need to say any more? Our approach is agnostic to which RL algorithm is used, so maybe we can simply say we use PPO and here's the basic idea -- the eqns won't be used in the rest of the paper.}

% \begin{align}
% \mathcal{L}_{\theta_{k}}^{\textrm{CLIP}}(\theta)=\underset{\tau \sim \pi_{k}}{\mathbb{E}}\left[\sum_{t=0}^{T}\left[\min \left(r_{t}(\theta) \hat{A}_{t}^{\pi_{k}}, \operatorname{clip}\left(r_{t}(\theta), 1-\epsilon, 1+\epsilon\right) \hat{A}_{t}^{\pi_{k}}\right)\right]\right]
% \end{align}

% where $r_{t} = \frac{\pi_\theta(a_t|s_t)}{\pi_{\theta_{old}}(a_t|s_t)}$ is the probability ratio between the old $\pi_{\theta_{\textrm{old}}}$ and current $\pi_{\theta_{\textrm{}}}$ policy. In addition, we employ the generalized advantage estimator, $ GAE(\gamma,\lambda)$ \cite{schulman2015high} to obtain an approximation of the advantage function defined as follows: 

% \begin{equation}
%     \hat{A}_{t} = \sum^\infty_{l=0}(\gamma \lambda)^l\delta_{t+l}^V
% \end{equation}

% where $\delta_{t}^V$ are the Bellman residuals.

\subsection{Agent Dynamics}\label{sec:dynamics}
Real robotic systems' inertia imposes limits on linear and angular acceleration. Thus, we assume a second-order unicycle model for the robot \cite{lavalle2006planning}:
\begin{equation}\label{eq:dynamics_model}
    \begin{array}{ll}
    \dot{x}=v\cos{\psi} & \dot{v}=u_a \\
    \dot{y}=v\sin{\psi} & \dot{\omega}=u_\alpha \\
    \dot{\psi}=\omega 
    \end{array}
\end{equation}
where $x$ and $y$ are the agent position coordinates and $\psi$ is the heading angle in a global frame. $v$ is the agent forward velocity, $\omega$ denotes the angular velocity and, $u_a$ the linear and $u_\alpha$ angular acceleration, respectively. %Hence, the agents control input vector is $\vu=[a,\alpha]$ and agents control state $\vx=[x,y,\theta,v,w]$.}

\subsection{Modeling Other Agents' Behaviors}\label{sec:others}

In a real scenario, agents may follow different policies and show different levels of cooperation. Hence, in contrast to previous approaches, we do not consider all the agents to follow the same policy \cite{everett2019collision,fan2018fully}. At the beginning of an episode, each non-ego agent either follows a cooperative or a non-cooperative policy. For the cooperative policy, we employ the Reciprocal Velocity Obstacle (RVO) \cite{van2008reciprocal} model with a random cooperation coefficient\footnote{This coefficient is denoted as $\alpha_A^B$ in \cite{van2011reciprocal}} $c^i \sim \mathcal{N}(0.1,1)$ sampled at the beginning of the episode. The ``reciprocal'' in RVO means that all agents follow the same policy and use the cooperation coefficient to split the collision avoidance effort among the agents (e.g., a coefficient of 0.5 means that each agent will apply half of the effort to avoid the other). In this work, for the non-cooperative agents, we consider both constant velocity (CV) and non-CV policies. The agents following a CV model drive straight in the direction of their goal position with constant velocity. The agents following a non-CV policy either move in sinusoids towards their final goal position or circular motion around their initial position. %Note that RVO agents with low cooperation coefficient (e.g. $c_i \le 0.5$) can also be considered as non-cooperative.

\section{METHOD}\label{sec:method}
Learning a sequence of intermediate goal states that lead an agent toward a final goal destination can be formulated as a single-agent sequential decision making problem. Because parts of the environment can be difficult to model explicitly, the problem can be solved with a reinforcement learning framework.
Hence, we propose a two-level planning architecture, as depicted in Figure \ref{fig:architecture_proposed}, consisting of a subgoal recommender (Section \ref{sec:rl}) and an optimization-based motion planner (Section \ref{sec:others}). We start by defining the RL framework and our's policy architecture (Section\ref{sec:rl}). Then, we formulate the MPC to execute the policy's actions and ensure local collision avoidance (Section\ref{sec:lmpcc}). %the agent's dynamics model (Section\ref{sec:dynamics}) followed by the description of the surrounding agents' policies (Section\ref{sec:lmpcc}). 

\subsection{Learning a Subgoal Recommender Policy}\label{sec:rl}

We aim to develop a decision-making algorithm to provide an estimate of the cost-to-go in dynamic environments with mixed-agents. In this paper, we propose to learn a policy directly informing which actions lead to higher rewards.

\subsubsection{RL Formulation}
As in \cite{chen2017decentralized}, the observation vector is composed by the ego-agent and the surrounding agents states, defined as:
\begin{equation}\begin{aligned}
\mathbf{s}_t &=\left[d_\vg,\mathbf{p}_t-\mathbf{g}, v_{\textrm{ref}}, \psi, r\right] \,\,\, \textrm{(Ego-agent)} \\
\mathbf{s}^{i}_t &=\left[\mathbf{p}^i_t, \mathbf{v}^i_t, r^i, d^ i_t, r^i+r\right] \,\, \forall i \in \{1,n\} \,\,\, \textrm{(Other agents)}
\end{aligned}\end{equation}
where $\vs_t$ is the ego-agent state and $\vs^i_t$ the $i$-th agent state at step $t$. Moreover, $d_\vg = \norm{\mathbf{p}_t-\mathbf{g}}$ is the ego-agent's distance to goal and $d^i_t = \norm{\mathbf{p}_t-\mathbf{p}^i_{t}}$ is the distance to the $i$-th agent. %At each step $k$, the observations is defined as the joint-state of the ego agent and other agents, $\vo_k = [\mathbf{s}_k,\mathbf{S}_k]$.

Here, we seek to learn the optimal policy for the ego-agent $\pi:(\mathbf{s}_t,\vS_t)\xrightarrow{}\va_t$ 
%\meXX{should use different letter than $\pi$ or somehow denote that this $\pi$ is for cost-to-go?} explain variables and remove uppscript i} 
mapping the ego-agent's observation of the environment to a probability distribution of actions. We consider a continuous action space $\mathcal{A} \subset \R^2$ and define an action as position increments providing the direction maximizing the ego-agent rewards, defined as:
\begin{subequations}
\begin{align}
    \mathbf{p}_{t}^\textrm{ref} = \mathbf{p}_t + \bm{\delta}_t \\
    \pi_{\theta^\pi}(\mathbf{s}_t,\mathbf{S}_t) = \bm{\delta}_{t} =  [\bm{\delta}_{t,x},\bm{\delta}_{t,y}] \label{eq:subgoal_policy}\\
    \norm{\bm{\delta}_{t}} \le N v_{\textrm{max}},
\end{align}
\end{subequations}
where $\bm{\delta}_{k,x},\bm{\delta}_{k,y}$ are the $(x,\,y)$ position increments, $v_{\textrm{max}}$ the maximum linear velocity and $\theta^\pi$ are the network policy parameters. Moreover, to ensure that the next sub-goal position is within the planning horizon of the ego-agent, we bound the action space according with the planning horizon $N$ of the optimization-based planner and its dynamic constraints, as represented in \cref{eq:subgoal_policy}.

We design the reward function to motivate the ego-agent to reach the goal position while penalizing collisions:
\begin{equation}
\label{eq:reward}
R\left(\mathbf{s}, \mathbf{a}\right)=\left\{\begin{array}{ll}
r_{\textrm{goal}} & \text { if } \mathbf{p}=\mathbf{p}_{g} \\
r_{\textrm{collision}} & \text { if } d_{\min }<r+r^i \,\,\, \forall i \in \{1,n\} \\
r_t & \text { otherwise }
\end{array}\right.
\end{equation}
where $d_{min} = \min\limits_{i} \norm{\vp - \vp^ i}$  is the distance to the closest surrounding agent. $r_t$ allows to adapt the reward function as shown in the ablation study (Sec.\ref{sec:reward}), %to dense, sparse or discrete rewards. 
$r_{\textrm{goal}}$ rewards the agent if reaches the goal $r_{\textrm{collision}}$ penalizes if it collides with any other agents.  In Section. \ref{sec:reward} we analyze its influence in the behavior of the learned policy. 

\subsubsection{Policy Network Architecture}\label{sec:rl}

%@Michael: Should we have this? This is the same as you did except for the hidden state part. What do you think?}
%\meXX{Yeah it seems good to mention this}

\begin{figure}[!t]
    \centering
    \includegraphics[width=\columnwidth,trim = 0 0 0 0, clip]{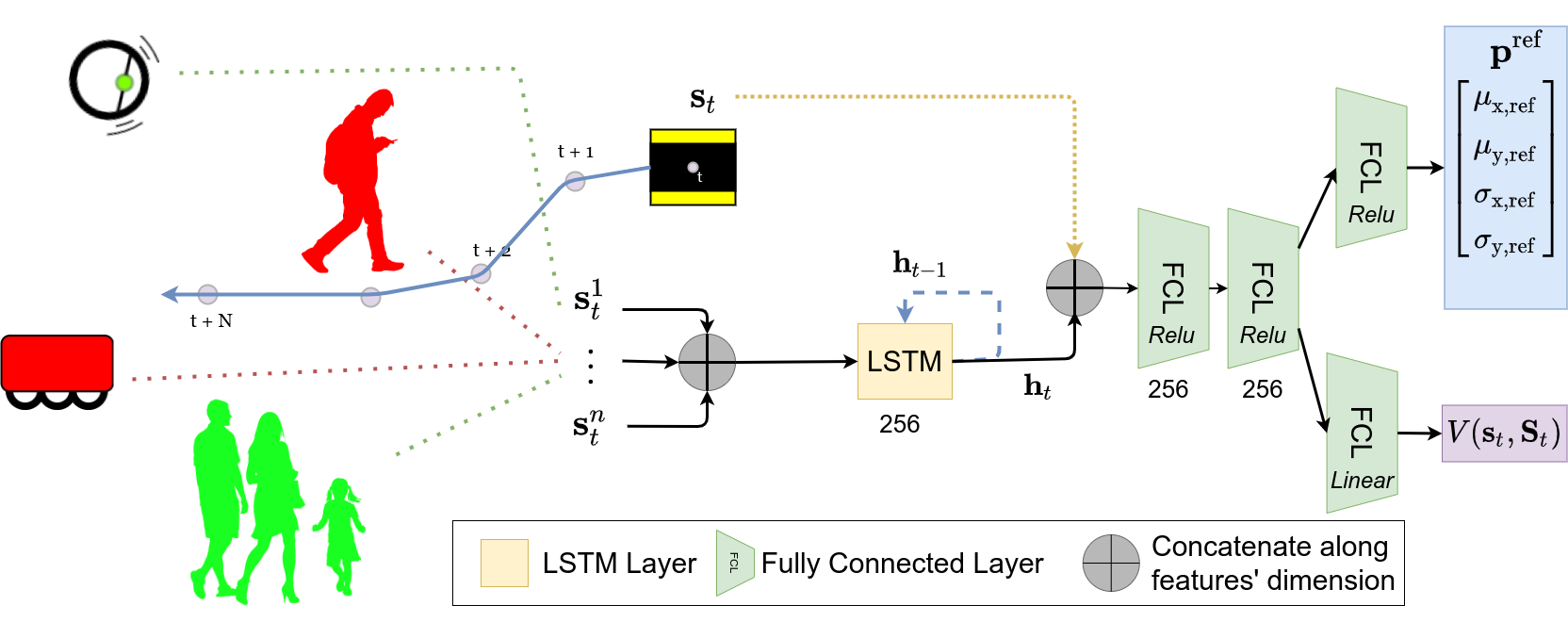}
    \caption{Proposed network policy architecture.}
    \label{fig:network_architecture}
\end{figure}

A key challenge in collision avoidance among pedestrians is that the number of nearby agents can vary between timesteps.
Because feed-forward NNs require a fixed input vector size, prior work~\cite{everett2019collision} proposed the use of Recurrent Neural Networks (RNNs) to compress the $n$ agent states into a fixed size vector at each time-step.
Yet, that approach discarded time-dependencies of successive observations (i.e., hidden states of recurrent cells).

Here, we use the ``store-state'' strategy, as proposed in \cite{kapturowski2018recurrent}. During the rollout phase, at each time-step we store the hidden-state of the RNN together with the current state and other agents state, immediate reward and next state, $(\vs_k, \vS_k,\vh_k,\vr_k,\vs_{k+1})$. Moreover, the previous hidden-state is feed back to warm-start the RNN in the next step, as depicted in Fig.\ref{fig:network_architecture}. During the training phase, we use the stored hidden-states to initialize the network. Our policy architecture is depicted in \cref{fig:network_architecture}. We employ a RNN to encode a variable sequence of the other agents states $\vS_k$ and model the existing time-dependencies. Then, we concatenate the fixed-length representation of the other agent's states with the ego-agent's state to create a join state representation. This representation vector is fed to two fully-connected layers (FCL). The network has two output heads: one estimates the probability distribution parameters $\pi_{\theta^\pi}(\vs,\vS)  \sim \cN(\mu,\sigma)$ of the policy's action space and the other estimates the state-value function $V^{\pi}\left(s_{t}\right):=\mathbb{E}_{s_{t+1: \infty},}\left[\sum_{l=0}^{\infty} r_{t+l}\right]$. $\mu$ and $\sigma$ are the mean and variance of the policy's distribution, respectively.

\subsection{Local Collision Avoidance: Model Predictive Control}\label{sec:lmpcc} 
%Recently, there is an increased interest on MPC for RL \cite{bellegarda2019combining,farshidian2019deep,lowrey2018plan}~\meXX{i'd remove this sentence}. 
Here, we employ MPC to generate locally optimal commands respecting the kino-dynamics and collision avoidance constraints. To simplify the notation used, hereafter, we assume the current time-step $t$ as zero. %Moreover, we feedback information about the feasibility of the MPC solution to the RL agent. Hence, the RL agent can use this information to proper evaluate the cost-to-go provided to the MPC.

\subsubsection{State and Control Inputs}
We define the ego-agent control input vector as $\vu=[u_a,u_\alpha]$ and the control state as $\vx=[x,y,\psi,v,w] \in \mathbb{R}^ 5$ following the dynamics model defined in \cref{sec:dynamics}.
%We define the ego-agent control vector $\vu=[a,\alpha] \in \mathbb{R}^ 2$, where $a$ is the linear acceleration and $\alpha$ the angular acceleration . Moreover, the ego-agent control state $\vx$ is composed by the the ego-agent x-y positions and orientation $\theta$ in a global frame and, the forward $v$ and angular velocity $\omega$. Hence, $\vx=[p_x,p_y,\theta,v,w] \in \mathbb{R}^ 5$.
%\meXX{much of this seems repeated from the vehicle model. also, do we want to use $\mathbf{p}_x$ instead of $x$?} Should I just remove it? or just define $\mathbf{u}$ and $\mathbf{x}$ here?}

\subsubsection{Dynamic Collision Avoidance}

We define a set of nonlinear constraints to ensure that the MPC generates collision-free control commands for the ego-agent (if a feasible solution exists). To limit the problem complexity and ensure to find a solution in real-time, we consider a limited number of surrounding agents $\cX^m$, with $m \leq n$. Consider $\cX^n=\{\vx_1,\dots,\vx_n\} $ as the set of all surrounding agent states, than %$\cS^m=\{\forall \vs_j \in \cS^m, \exists \vs_i \in \cS^n: d(\vs_j,\vs) \leq d(\vs_i,\vs) \}$.
the set of the $m$-th closest agents is:
\begin{definition}[]\label{def:set_close_agents}
    A set $\cX^m \subseteq \cX^n$ is the set of the $m$-th closest agents if the euclidean distance $\forall \vx_j \in \cX^m, \forall \vx_i \in \cX^n\setminus\cX^m: \norm{\vx_j,\vx} \leq \norm{\vx_i,\vx}$.
\end{definition}

We represent the area occupied by each agent $\pazocal{O}^i$ as a circle with radius $r_i$. 
%is given by $\pazocal{O}_k^{\textrm{i}} = \bigcup_{i \in \{1,\dots,n \}} \A_i(\vs_i(t))$, where $\boldsymbol{z}_i(t)$ denotes the state of moving obstacle $i$ at time $t$. 
To ensure collision-free motions, we impose that each circle $i \in \{1,\dots,n\}$ $i$ does not intersect with the area occupied by the ego-agent resulting in the following set of inequality constraints:
\begin{equation}\label{eq:coll_constraint}
\begin{split}
 \!\!c_k^{i}(\vx_k,\vx_k^i) = \norm{\vp_k,\vp^i_k} \geq r+r_i, \hspace{5mm} %\forall i \in \{1,\dots,n\}
%\Bigg\rvert_{k,j}
\end{split}
\end{equation}
for each planning step $k$.
This formulation can be extended for agents with general quadratic shapes, as in \cite{brito2019model}.

\subsubsection{Cost function}\label{sec:cost_function}
The subgoal recommender provides a reference position $\vp^{\textrm{ref}}_0$ guiding the ego-agent toward the final goal position $\vg$ and minimizing the cost-to-go while accounting for the other agents. The terminal cost is defined as the normalized distance between the ego-agent's terminal position (after a planning horizon $N$) and the reference position (with weight coefficient $Q_N$): 
\begin{align}
    J_N(\vp_N,\pi(\vx,\vX)) =  \norm{\frac{\vp_{N} - \vp^{\textrm{ref}}_0}{\vp_{0} - \vp^{\textrm{ref}}_0}}_{Q_N},
\end{align}
%where .
%\meXX{the arguments to $J_N$ are a little weird, since you just use $p$ terms}
%true. $p_N$ comes from $\vx$ and $p_{ref}$ from $\pi$. Should I just replace $\vx$ for $p_N$?.}
%\meXX{yes, lets keep this eqn simple, just using $p$, and then somewhere we say that we say that we set $p=\pi(\ldots)$ in our cascaded system.}
%
To ensure smooth trajectories, we define the stage cost as a quadratic penalty on the ego-agent control commands
\begin{align}
    J_{k}^{\vu}(\vu_k) = \norm{\vu_k}_{Q_{u}},\quad k = \{0, 1, \dots, N-1\},
\end{align}
where $Q_{u}$ is the weight coefficient.

%To introduce some social clearance between the ego-agent and the others, we introduce an additional cost term
%\begin{equation}\label{eq:repulsive}
%J_{\text{social}}(\boldsymbol{z}_k) = Q_R \sum^{n}_{i = 1} \mathcal{N}(\Delta\mathbf{p},\mu,\Sigma)
%\end{equation}
% where $\Delta\mathbf{p}=\mathbf{p}^i-\mathbf{p}^0$ is the relative position vector from the $i$-th agent to the ego-agent, $\mathcal{N}$ is a bivariate Gaussian distribution with zero mean ,$\mu=[0,0]$, and variance proportional to the relative velocity $\Sigma=\begin{bmatrix}||v^i_ x-v^0_ x|| & 0 \\ 0 & ||v^i_ y-v^0_ y||\end{bmatrix}$ between the agents. Finally, $\{Q_{x},Q_{y},Q_u,Q_R\}$ are the design weights.

\subsubsection{MPC Formulation}
The MPC is then defined as a non-convex optimization problem
\begin{equation}\label{eq:control_problem}
    \begin{aligned}
    \min\limits_{\vx_{1:N}, \vu_{0:N-1}}  ~~        
                            & J_N(\vx_N,\vp^{\textrm{ref}}_0) + \sum_{k=0}^{N-1} J^{u}_k(\vu_k) \\
    \text{s.t.}	~~	    & \vx_0 = \vx(0), \quad \textrm{\eqref{eq:state_control_constraints}, \eqref{eq:dynamics_model}}, \\
                            &   c_k^{i}(\vx_k,\vx_k^i)  > r+r_i ,\\ 
                            & \vu_{k} \in \cU, \quad
                            \vx_k \in \cS,\\
                            &\forall i \in \{1,\dots,n \}; \, \forall k\in \{0,\dots,N-1\}.
    \end{aligned}
\end{equation}
%Because we are optimizing over a prediction horizon, we need an estimate of the other agents' future positions. 
In this work, we assume a constant velocity model estimate of the other agents' future positions, as in \cite{brito2019model}.

\subsection{PPO-MPC}
In this work, we train the policy using a state-of-art method, Proximal Policy Optimization (PPO) \cite{schulman2017proximal}, but the overall framework is agnostic to the specific RL training algorithm.
We propose to jointly train the guidance policy $\pi_{\theta^\pi}$ and value function $V_{\theta^V}(\vs)$ \textit{with} the MPC, as opposed to prior works~\cite{everett2019collision} that use an idealized low-level controller during policy training (that cannot be implemented on a real robot). \cref{alg:ppo-mpc} describes the proposed training strategy and has two main phases: supervised and RL training. %The usage of the MPC as supervisor removes the need a large set of training data \meXX{why? because the policy is already decent/doesn't need to learn collision avoidance?} because the supervised step that would be done with a dataset is not needed because for the first training steps we use the mpc as the right answer and do supervised training}. 
First, we randomly initialize the policy and value function parameters $\{\theta^\pi,\theta^V\}$. Then, at the beginning of each episode we randomly select the number of surrounding agents between $[1,n_{\textrm{agents}}]$, the training scenario and the surrounding agents policy. More details about the different training scenarios and $n_{\textrm{agents}}$ considered is given in Sec.\ref{sec:training}.

%Although it may be trivial for an optimization-based planner to drive towards a goal position 
An initial RL policy is unlikely to lead an agent to a goal position. Hence, during the warm-start phase, we use the MPC as an expert and perform supervised training to train the policy and value function parameters for $n_{\textrm{MPC}}$ steps. By setting the MPC goal state as the ego-agent final goal state $\vp^{\textrm{ref}}=\vg$ and solving the MPC problem, we obtain a locally optimal sequence of control states $\vx^*_{1:N}$. For each step, we define $\va^*_t = \vx^*_{t,N}$ and store the tuple containing the network hidden-state, state, next state, and reward in a buffer $\cB \la \{\vs_k,\va^*_t,r_k,\vh_k,\vs_{k+1}\}$. %In contrast with previous approaches (\meXX{all previous approaches, or just our prior approaches/prior collision avoidance approaches}, not sure about all, lets discuss}), we store also the network hidden-state following a similar strategy to a recurrent experience replay \cite{kapturowski2018recurrent} \meXX{isnt this discussed earlier?}. 
Then, we compute advantage estimates~\cite{schulman2015high} and perform a supervised training step
\begin{eqnarray}
    \theta^V_{k+1} &\!\!\!\!\!=\!\!\!\!\!& \argmin_{\theta^V} \mathbb{E}_{(\va_k,\vs_k,r_k)\sim \cD_{\textrm{MPC}}}[\norm{V_\theta(\vs_k)-V_k^{\textrm{targ}}}] \label{eq:supervised} \\
    \theta^\pi_{k+1} &\!\!\!\!\!=\!\!\!\!\!& \argmin_\theta \mathbb{E}_{(\va^*_k,\vs_k)\sim \cD_{\textrm{MPC}}}[\norm{\va_k^*-\pi_\theta(\vs_k)}] \label{eq:supervised_2}
\end{eqnarray}
where $\theta^V,\theta^\pi$ are the value function and policy parameters, respectively. Note that $\theta^V$ and $\theta^\pi$ share the same parameter except for the final layer, as depicted in Fig.\ref{fig:network_architecture}.
Afterwards, we use Proximal Policy Optimization (PPO) \cite{schulman2017proximal} with clipped gradients for training the policy. PPO is a on-policy method addressing the high-variance issue of policy gradient methods for continuous control problems. We refer the reader to \cite{schulman2017proximal} for more details about the method's equations. Please note that our approach is agnostic to which RL algorithm we use. Moreover, to increase the learning speed during training, we gradually increase the number of agents in the training environments (curriculum learning \cite{bengio2009curriculum}).

\begin{algorithm}[t]
\caption{PPO-MPC Training}\label{alg:ppo-mpc}
\begin{algorithmic}[1]
\State \textbf{Inputs:} planning horizon $H$, value fn. and policy parameters $\{\theta^V,\theta^\pi\}$, number of supervised and RL training episodes $\{n_{\textrm{MPC}}, n_{\textrm{episodes}}\}$, number of agents $n$, $n_{\textrm{mini-batch}}$, and  reward function $R(\vs_t,\va_t,\va_{t+1})$
 
 %\State Randomly initialize all agents initial and goal state: $\{\vs^0_0,\dots,\vs^n_0\}\sim \cS $ , $\{\vg^0,\dots,\vg^n\}\sim \cS $ 
\State Initialize states: $\{\vs^0_0,\dots,\vs^n_0\}\sim \cS $ , $\{\vg^0,\dots,\vg^n\}\sim \cS $
\While{$episode < n_{\textrm{episodes}}$}
        % Supervised Training
    \State Initialize $\cB \la \emptyset$ and $h_0 \la \emptyset$
    \For{$k=0,\dots,n_{\textrm{mini-batch}}$}
        \If{$episode \le n_{\textrm{MPC}}$}
                \State Solve Eq.\ref{eq:control_problem} considering $\vp^{\textrm{ref}} = \vg$ %to compute the control state sequence $\vx^*_{1:N}$
                \State Set $\va^*_t=\vx^*_N$
        \Else{}
            \State $\vp^{\textrm{ref}}=\pi_\theta(\vs_t,\vS_t)$       
        \EndIf
        \State ${\{\vs_k,\va_k,r_k,\vh_{k+1},\vs_{k+1},\textrm{done}\} = \textrm{Step}(\vs^*_t,\va^*_t,\vh_t)}$
        \State Store $\cB \la \{\vs_k,\va_k,r_k,\vh_{k+1},\vs_{k+1},\textrm{done}\}$
        \If{done}
            \State $episode\ +=1$
            \State Reset hidden-state: $h_t \la \emptyset$
             \State Initialize: ${\{\vs^0_0,\dots,\vs^n_0\}\sim \cS, \{\vg^0,\dots,\vg^n\}\sim \cS}$ 
        \EndIf
    \EndFor
    \If{$episode \le n_{\textrm{MPC}}$}
        \State Supervised training: Eq.\ref{eq:supervised} and Eq.\ref{eq:supervised_2}
    \Else{}
        \State PPO training \cite{schulman2017proximal} 
    \EndIf
\EndWhile

\State \Return $\{\theta^V,\theta^\pi\}$
%     \If{$\theta_k>s_m$}
%         \State $m=m+1$
%     \EndIf 
%     \State $k=k+1$
% \EndWhile
\end{algorithmic}
\end{algorithm}

\section{RESULTS}\label{sec:results} 

This section quantifies the performance throughout the training procedure, provides an ablation study, and compares the proposed method (sample trajectories and numerically) against the following baseline approaches:
\begin{itemize}
    \item MPC: Model Predictive Controller from \cref{sec:lmpcc} with final goal position as position reference, $\vp_{\textrm{ref}}=\vg$;
    \item DRL~\cite{everett2019collision}:  state-of-the-art Deep Reinforcement Learning approach for multi-agent collision avoidance %trained on first-order unicycle model \cite{lavalle2006planning}};
    %\item DRL-2 ~\cite{everett2019collision}: state-of-the-art Deep Reinforcement Learning approach for multi-agent collision avoidance trained on second-order unicycle model, as described in Eq.\ref{eq:dynamics_model}}.
\end{itemize}
To analyze the impact of a realistic kinematic model during training, we consider two variants of the DRL method ~\cite{everett2019collision}: the same RL algorithm~\cite{everett2019collision} was used to train a policy under a first-order unicycle model, referred to as DRL, and a second-order unicycle model (Eq.\ref{eq:dynamics_model}), referred to as DRL-2. %All methods are evaluated with second-order unicycle model (Eq.\ref{eq:dynamics_model}).
All experiments use a second-order unicycle model (Eq.\ref{eq:dynamics_model}) in environments with cooperative and non-cooperative agents to represent realistic robot/pedestrian behavior.
Animations of sample trajectories accompany the paper.

\subsection{Experimental setup}\label{sec:experiment_setup}
%We implemented the described policy network architecture using TensorFlow \cite{abadi2016tensorflow}.
The proposed training algorithm builds upon the open-source PPO implementation provided in the Stable-Baselines \cite{stable-baselines} package. We used a laptop with an Intel Core i7 and 32~GB of RAM for training. To solve the non-linear and non-convex MPC problem of \cref{eq:control_problem}, we used the ForcesPro~\cite{FORCESPro} solver. If no feasible solution is found within the maximum number of iterations, then the robot decelerates. 
% Computation times discussion
All MPC methods used in this work consider collision constraints with up to the closest six agents so that the optimization problem can be solved in less than 20ms. Moreover, our policy's network has an average computation time of 2ms with a variance of 0.4ms for all experiments. Hyperparameter values are summarized in \cref{tab:hyperparameters}. 

%\newcolumntype{P}[1]{>{\centering\arraybackslash}p{#1}|}
\begin{table}[t]
\caption{Hyper-parameters. }
\label{tab:hyperparameters}
\centering
\begin{tabular}{|c|c||c|c|}
\hline
Planning Horizon $N$ & 2 s & Num. mini batches & 2048  \\ \hline
Number of Stages & 20 & $r_{\textrm{goal}}$ & 3 \\ \hline
$\gamma$ & 0.99 & $r_{\textrm{collision}}$ & -10 \\ \hline
Clip factor & 0.1 & Learning rate & $10^{-4}$ \\ \hline
\end{tabular}
\vspace{-0.2in}
\end{table}

\subsection{Training procedure}\label{sec:training}
\begin{figure}[t]
    \centering
    \includegraphics[width=0.8\columnwidth]{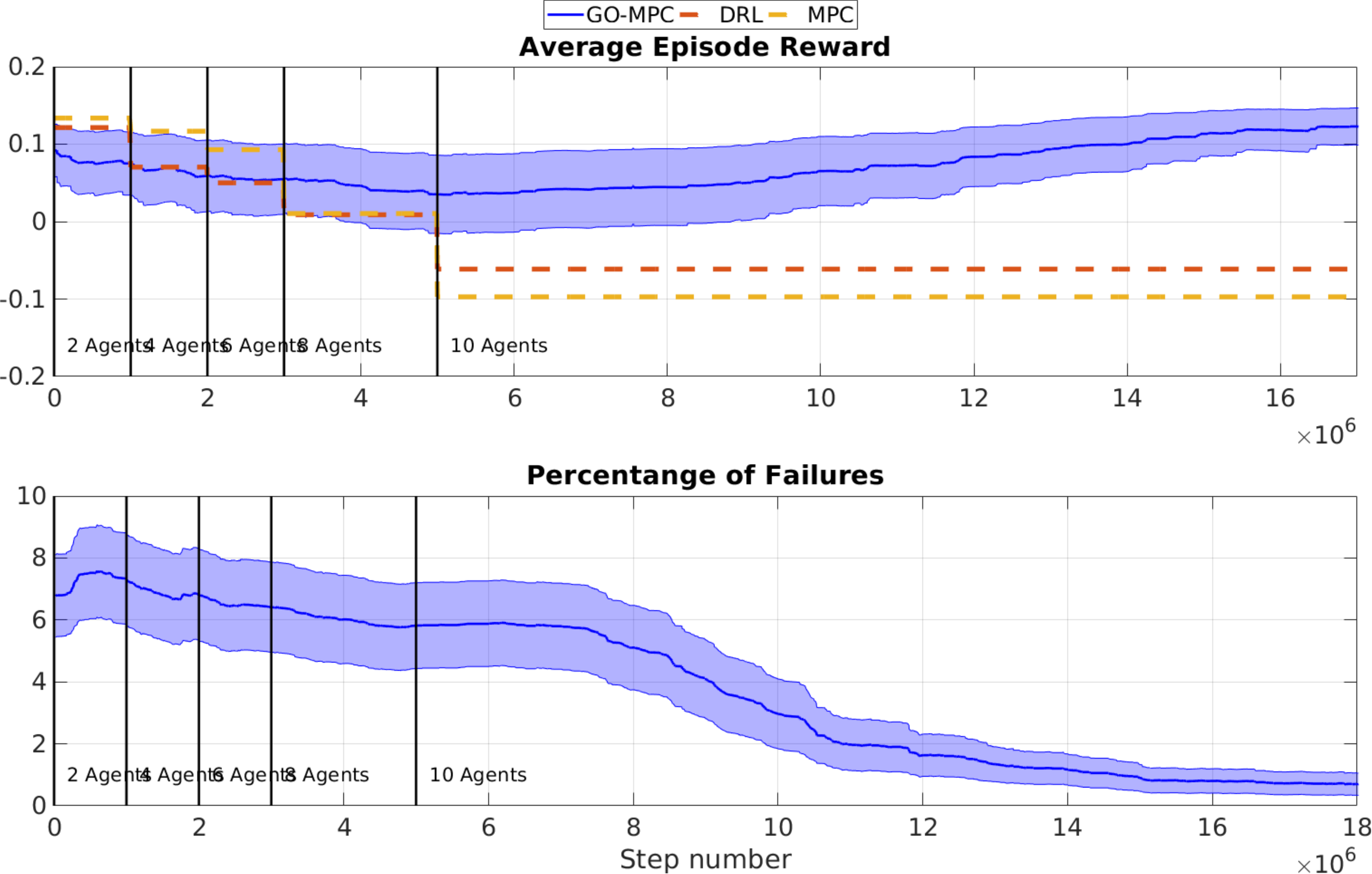}
    \caption{Moving average rewards and percentage of failure episodes during training. The top plot shows our method average episode reward vs DRL~\cite{everett2019collision} and simple MPC.}
    \label{fig:coll_evol_training}
    \vspace{-0.2in}
\end{figure}

To train and evaluate our method we have selected four navigation scenarios, similar to \cite{everett2019collision,chen2019crowd,fan2020distributed}:
\begin{itemize}
    \item \textbf{Symmetric swapping}: Each agent's position is randomly initialized in different quadrants of the $\mathbb{R}^2$ x-y plane, where all agents have the same distance to the origin. Each agent's goal is to swap positions with an agent from the opposite quadrant.
    \item \textbf{Asymmetric swapping}: As before, but all agents are located at different distances to the origin.
    %\item Each agent is assigned a position on a virtual circle centered on the plane $\mathbb{R}^2$. We sample a random circle radius $r_{\textrm{Circle}}$, , then for each pair of agents we randomly sample an angle $\psi\in [0,2\pi]$ and 
    \item \textbf{Pair-wise swapping}: Random initial positions; pairs of agents' goals are each other's intial positions
    \item \textbf{Random}: Random initial \& goal positions
\end{itemize}
Each training episode consists of a random number of agents and a random scenario.
At the start of each episode, each other agent's policy is sampled from a binomial distribution (80\% cooperative, 20\% non-cooperative). Moreover, for the cooperative agents we randomly sample a cooperation coefficient $c^i \sim \pazocal{U}(0.1,1)$ and for the non-cooperative agents is randomly assigned a CV or non-CV policy (i.e., sinusoid or circular).
\cref{fig:coll_evol_training} shows the evolution of the robot average reward and the percentage of failure episodes. 
The top sub-plot compares our method average reward with the two baseline methods: DRL (with pre-trained weights) and MPC. The average reward for the baseline methods (orange, yellow) drops as the number of agents increases (each vertical bar). In contrast, our method (blue) improves with training and eventually achieves higher average reward for 10-agent scenarios than baseline methods achieve for 2-agent scenarios. The bottom plot demonstrates that the percentage of collisions decreases throughout training despite the number of agents increasing.
%\meXX{Remove when you remove the bottom plot? $\Rightarrow$ The bottom sub-plot shows the number of successful episodes per rollout. As the number of agents increases, the number of successful events also decreases because the scenarios are more complex, requiring more steps to reach the goal. Nevertheless, around 10 million training steps, the number of successful events continuously increases while the number of collisions keeps going down, demonstrating that our method's performance keeps improving.}

\subsection{Ablation Study}\label{sec:reward}
A key design choice in RL is the reward function; here, we study the impact on policy performance of three variants of reward. The \textit{sparse} reward uses $r_t=0$ (only non-zero reward upon reaching goal/colliding). The \textit{time} reward uses $r_t=-0.01$ (penalize every step until reaching goal). The \textit{progress} reward uses $r_t = 0.01*(\norm{\vs_t-\vg}-\norm{\vs_{t+1}-\vg})$ (encourage motion toward goal). Aggregated results in Table \ref{tab:reward_performance} show that the resulting policy trained with a time reward function allows the robot to reach the goal with minimum time, to travel the smallest distance, and achieve the lowest percentage of failure cases. Based on these results, we selected the policy trained with the time reward function for the subsequent experiments.
%All experiments use the values $r_{\textrm{goal}}=3$ and $r_{\textrm{collision}}=-10$ for \cref{eq:reward}.
%\meXX{do these go in table 1 instead? what about $r_t$?}$r_t$ was previously defined in the text.}

%Secondly, we compared the training performance with and without collision constraints. The results show that having collision avoidance constraints during training increases significantly the training speed and policy performance. (This is still to be finished.)}

\begin{table*}[]
 \caption{Ablation Study: Discrete reward function leads to better policy than sparse, dense reward functions. Results are aggregated over 200 random scenarios with $n \in \{6,8,10\}$ agents. %\meXX{Better than sparse/dense/discrete would be sparse/time penalty/progress penalty?}
 }
        \centering
\begin{tabular}{|p{2cm}|p{1.2cm}|p{1.2cm}|p{1.2cm}|p{1.2cm}|p{1.2cm}|p{1.2cm}|p{1.2cm}|p{1.2cm}|p{1.2cm}|p{1.2cm}|p{1.2cm}|p{1.2cm}|}
\hline
\multicolumn{2}{|c|}{} & \multicolumn{3}{c|}{Time to Goal [s]} & \multicolumn{3}{c|}{\% failures (\% collisions / \% timeout) } & \multicolumn{3}{c|}{Traveled distance Mean [m]} \\ \hline
\multicolumn{2}{|c|}{\# agents}                    & \multicolumn{1}{c|}{6}                & \multicolumn{1}{c|}{8}       & \multicolumn{1}{c|}{10}                        & \multicolumn{1}{c|}{6}                &\multicolumn{1}{c|}{8}     & \multicolumn{1}{c|}{10}                       & \multicolumn{1}{c|}{6}                & \multicolumn{1}{c|}{8}       &\multicolumn{1}{c|}{10}        \\  \hline \hline
\multicolumn{2}{|l|}{Sparse Reward}                            & \multicolumn{1}{c|}{8.00}      &  \multicolumn{1}{c|}{8.51}          &     \multicolumn{1}{c|}{ 8.52}          & \multicolumn{1}{c|}{0 (0 / 0)}       & \multicolumn{1}{c|}{ 1 (0 / 1)}     &   \multicolumn{1}{c|}{2 (1 / 1) }            &  \multicolumn{1}{c|}{ 13.90}                &   \multicolumn{1}{c|}{14.34}       & \multicolumn{1}{c|}{14.31}   \\ \hline
\multicolumn{2}{|l|}{Progress Reward}                      &  \multicolumn{1}{c|}{8.9}     &  \multicolumn{1}{c|}{8.79}       &    \multicolumn{1}{c|}{9.01 }         & \multicolumn{1}{c|}{ 2 (1 / 1)}    &  \multicolumn{1}{c|}{3 (3 / 0)}        &  \multicolumn{1}{c|}{1 (1 / 0)  }        &  \multicolumn{1}{c|}{14.75}   &   \multicolumn{1}{c|}{14.57}       & \multicolumn{1}{c|}{14.63}    \\ \hline
\multicolumn{2}{|l|}{Time Reward}                &   \multicolumn{1}{c|}{\textbf{7.69}}              &   \multicolumn{1}{c|}{\textbf{8.03}}            &  \multicolumn{1}{c|}{\textbf{8.12}}       & \multicolumn{1}{c|}{\textbf{  0 (0 / 0) }}    &  \multicolumn{1}{c|}{\textbf{0 (0 / 0)   }}        & \multicolumn{1}{c|}{\textbf{   0 (0 / 0)   }}      & \multicolumn{1}{c|}{\textbf{13.25}}                &\multicolumn{1}{c|}{\textbf{14.01} }    &      \multicolumn{1}{c|}{\textbf{14.06}}         \\ 
\hline
\end{tabular}
 \label{tab:reward_performance}
 \vspace{-0.2in}
\end{table*}

\subsection{Qualitative Analysis}

\begin{figure}[!t]
	\centering
	\includegraphics[width=0.8\linewidth]{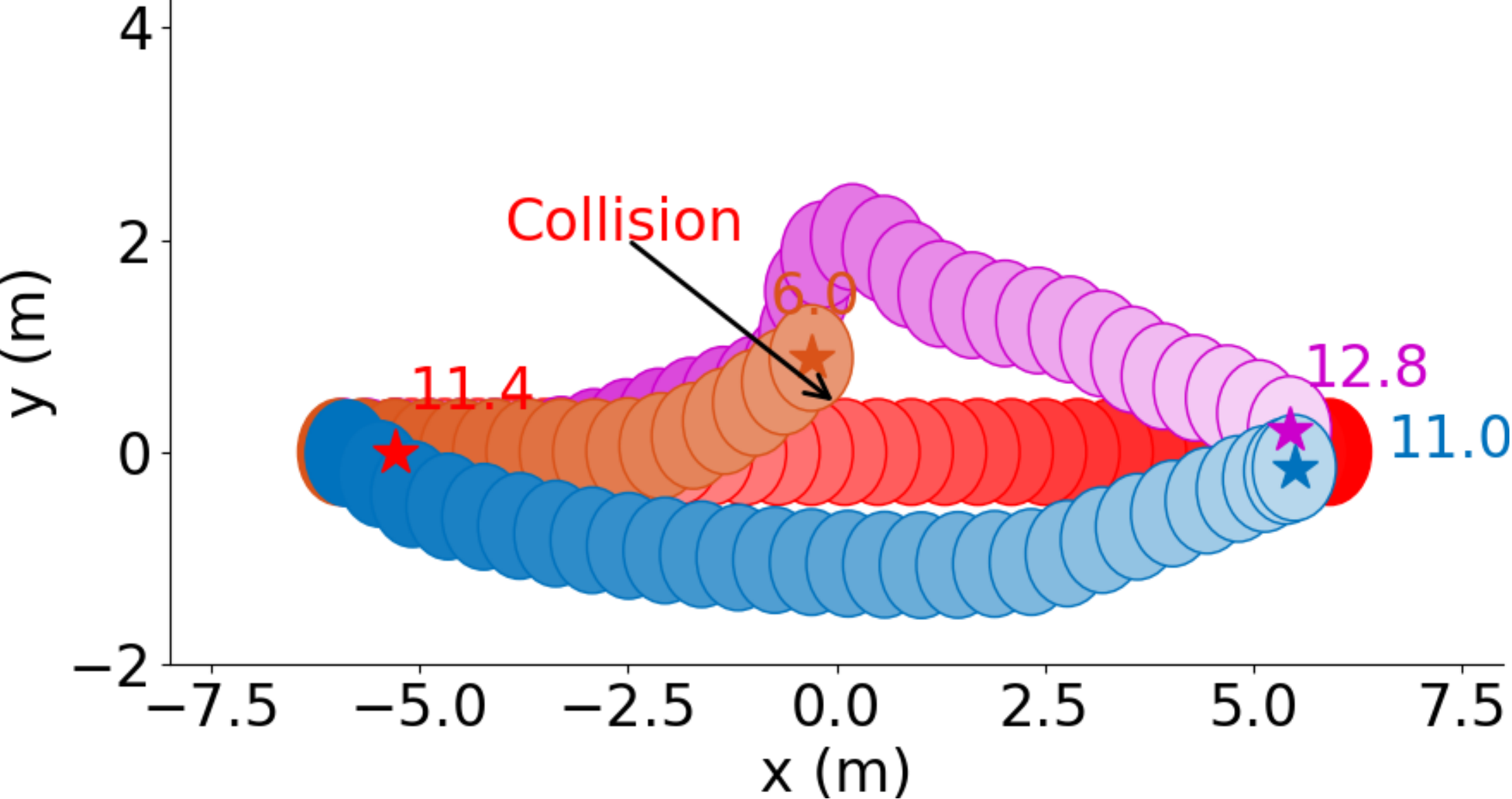}
	\caption{Two agents swapping scenario. In blue is depicted the trajectory of robot, in red the non-cooperative agent, in purple the DRL agent and, in orange the MPC.}
	\label{fig:2_agents_swap}
	\vspace{-0.1in}
\end{figure}

This section compares and analyzes trajectories for different scenarios. \cref{fig:2_agents_swap} shows that our method resolves a failure mode of both RL and MPC baselines.
The robot has to swap position with a non-cooperative agent (red, moving right-to-left) and avoid a collision. We overlap the trajectories (moving left-to-right) performed by the robot following our method (blue) versus the baseline policies (orange, magenta). The MPC policy (orange) causes a collision due to the dynamic constraints and limited planning horizon. The DRL policy avoids the non-cooperative agent, but due to its reactive nature, only avoids the non-cooperative agent when very close, resulting in larger travel time. Finally, when using our approach, the robot initiates a collision avoidance maneuver early enough to lead to a smooth trajectory and faster arrival at the goal.

\begin{figure}
    \centering
    \begin{subfigure}{0.31\linewidth}
        \centering
    	\includegraphics[width=\textwidth, trim = 0 0 0 0, clip]{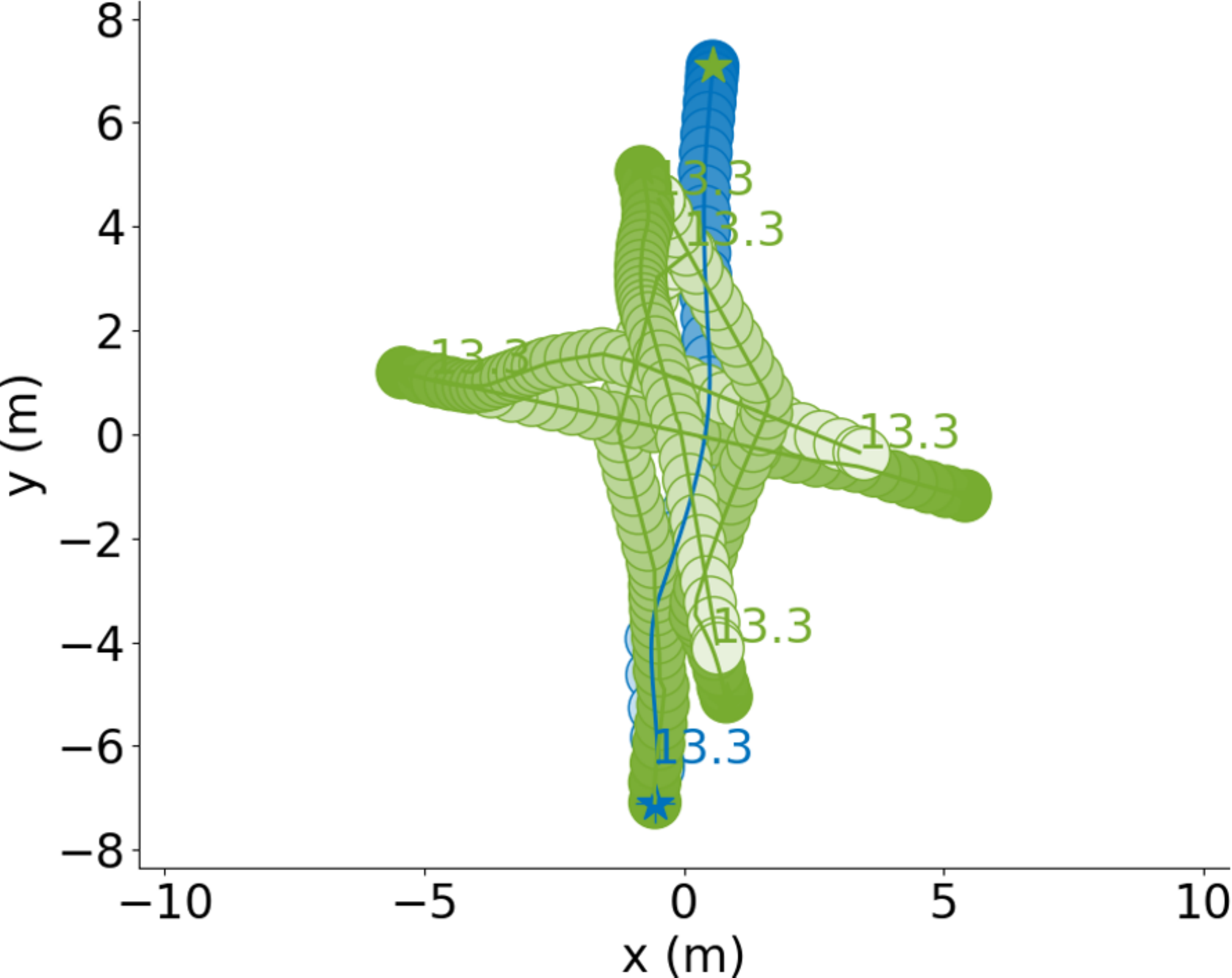}
    	\captionsetup{justification=centering}
    	\caption{}
    	%\caption{GO-MPC with 5 agents \\ (all coop.) \\}
    	\label{fig:mix_agents_cooperative}
    \end{subfigure}%
    %\vspace{1mm}
    \hspace{1mm}
    \begin{subfigure}{0.31\linewidth}
        \centering
    	\includegraphics[width=\textwidth, trim = 0 0 0 0, clip]{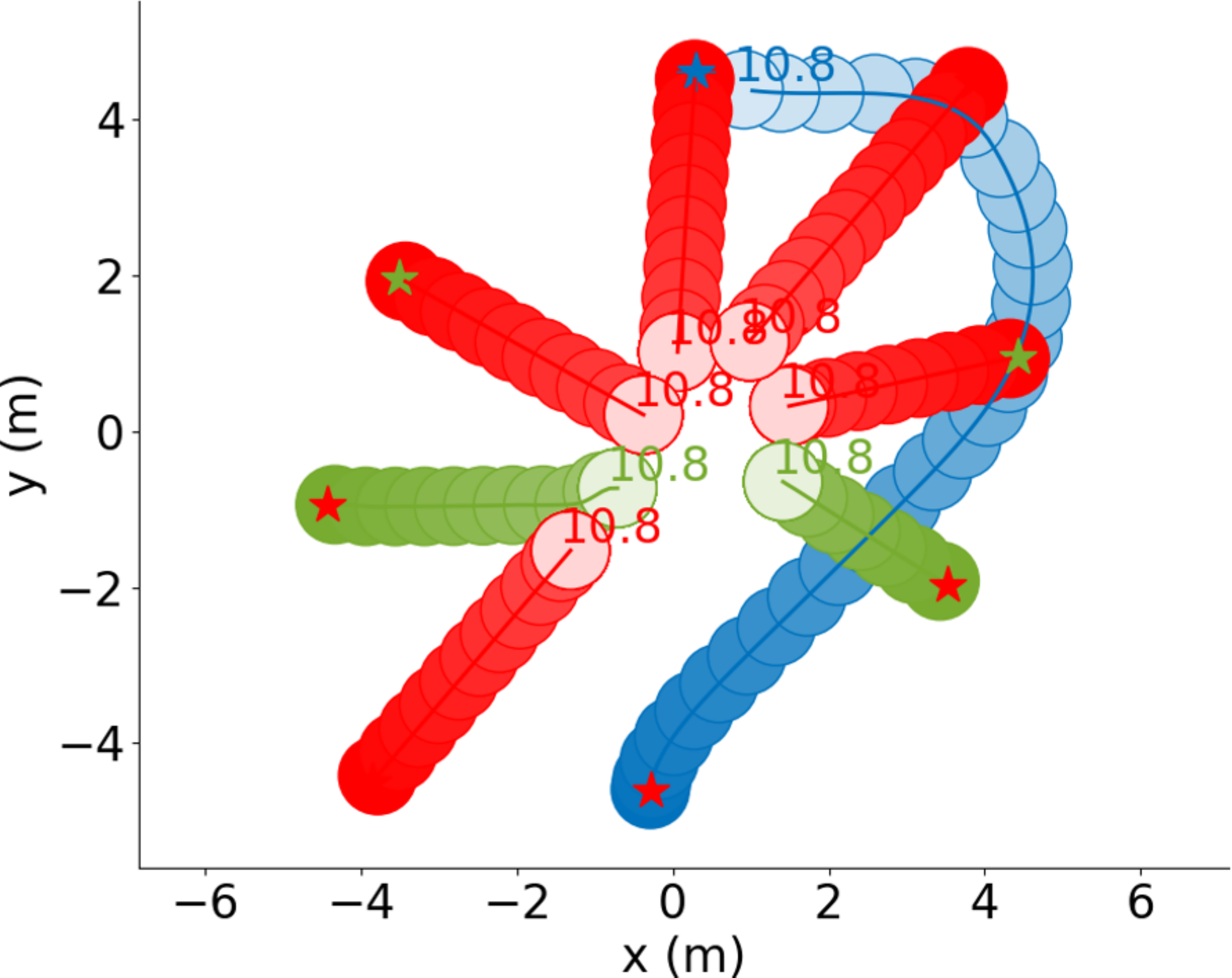}
    	\captionsetup{justification=centering}
    	\caption{}
    	%\caption{GO-MPC with 7 agents \\ (2 coop., 5 non-coop.)}
    	\label{fig:mix_agents_mix}
    \end{subfigure}%
    \begin{subfigure}{0.31\linewidth}
        \centering
    	\includegraphics[width=\textwidth, trim = 0 0 0 0, clip]{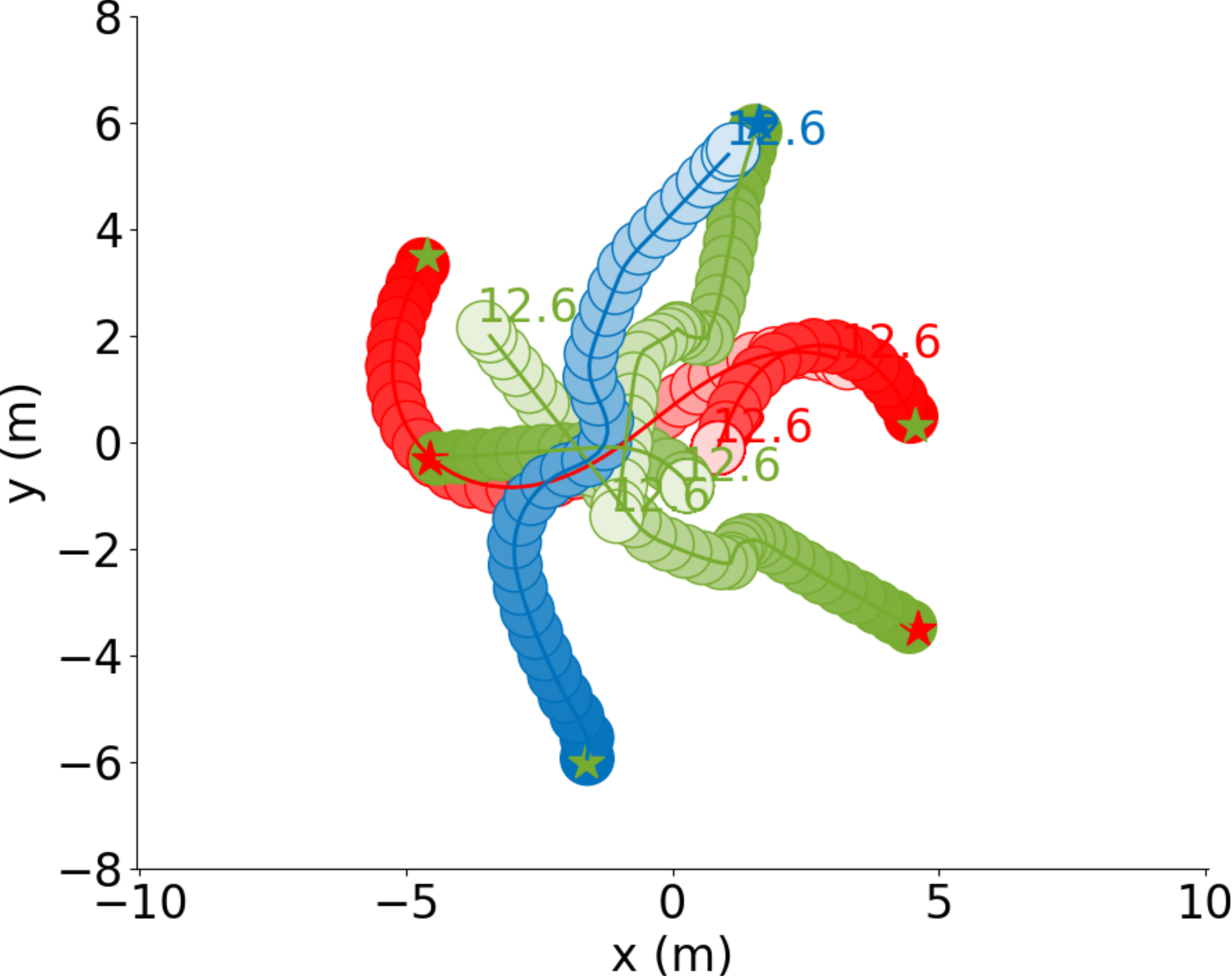}
    	\captionsetup{justification=centering}
    	\caption{}
    	%5\caption{GO-MPC with 5 agents \\ (3 coop., 2 non-coop.).}
    	\label{fig:non_cv_aggents}
    \end{subfigure}%
    \caption{Sample trajectories with mixed agent policies (robot: blue, cooperative: green, non-cooperative: red). In (a), all agents are cooperative; in (b), two are cooperative and five non-cooperative (const. vel.); in (c), three are cooperative and two non-cooperative (sinusoidal). The GO-MPC agent avoids non-cooperative agents differently than cooperative agents.}
    \label{fig:mix_agents}
    \vspace{-0.1in}
\end{figure}

\begin{figure}

    \centering
    \begin{subfigure}{0.33\linewidth}
        \centering
    	\includegraphics[width=\textwidth, trim = 0 0 0 0, clip]{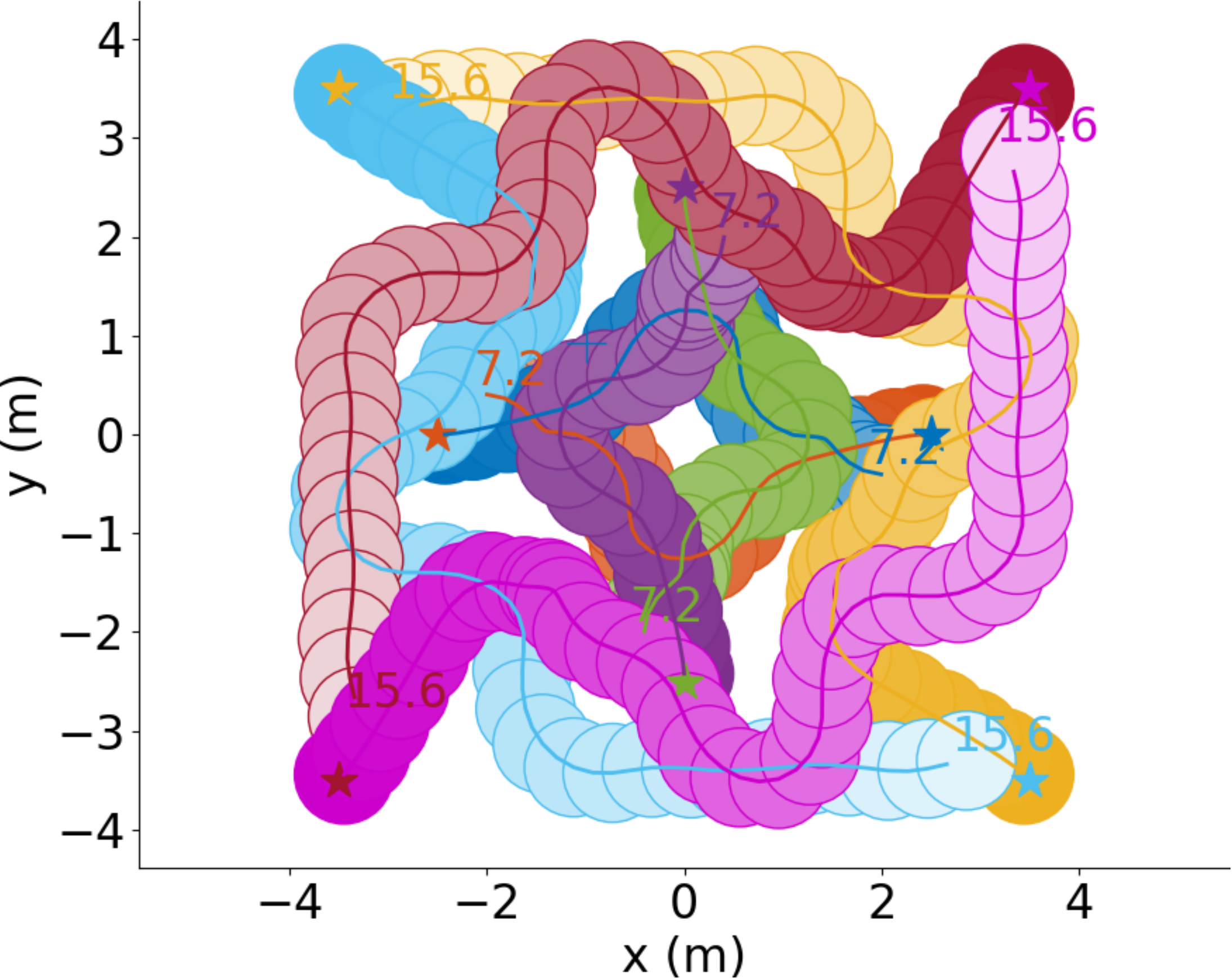}
    	\caption{DRL \cite{everett2019collision}}
    	\label{fig:homo_agents_rl}
    \end{subfigure}%
    \begin{subfigure}{0.33\linewidth}
        \centering
    	\includegraphics[width=\textwidth, trim = 0 0 0 0, clip]{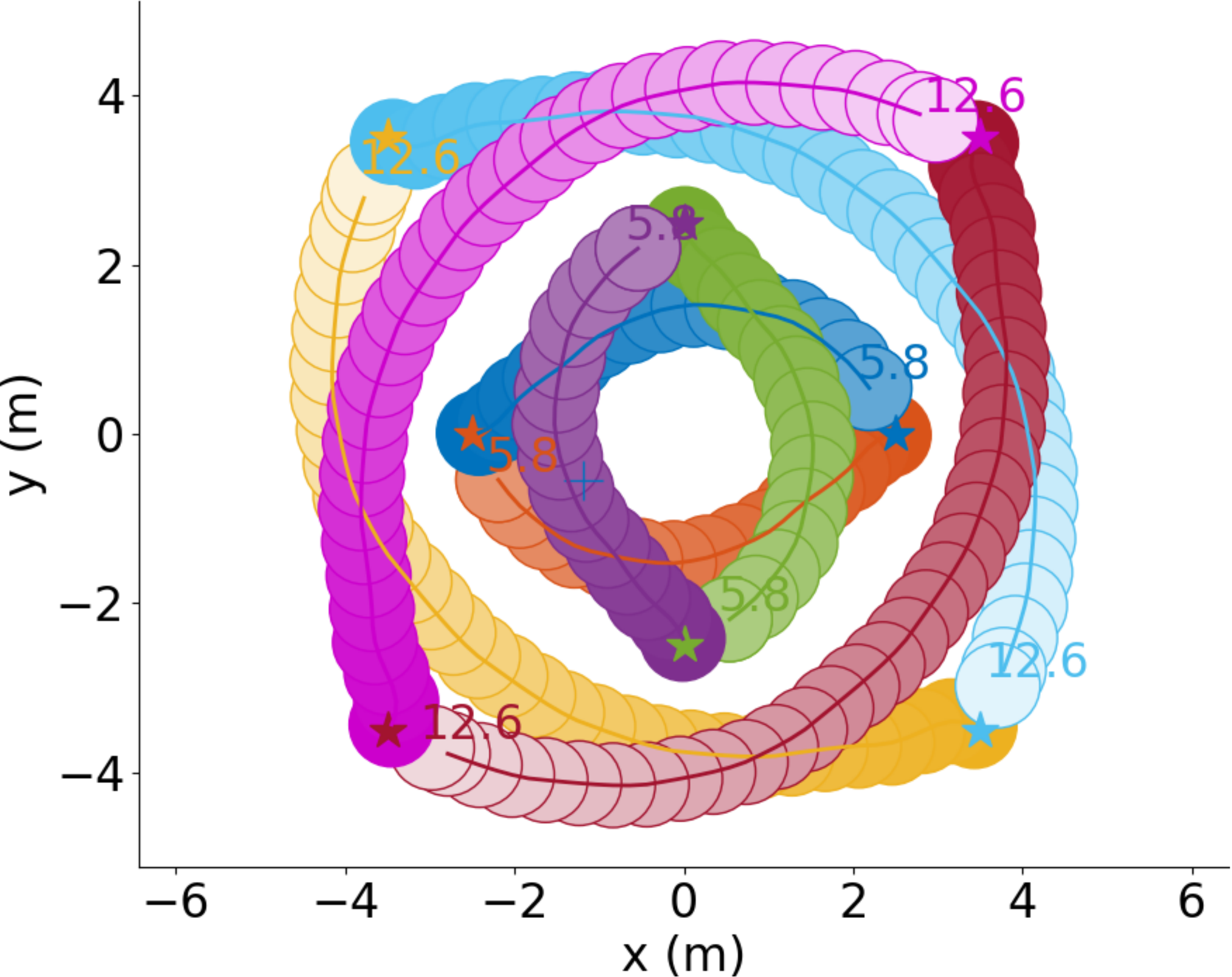}
    	\caption{\footnotesize{DRL-2 (ext. of \cite{everett2019collision})}}
    	\label{fig:homo_agents_rl_d2}
    \end{subfigure}%
    \begin{subfigure}{0.33\linewidth}
        \centering
    	\includegraphics[width=\textwidth, trim = 0 0 0 0, clip]{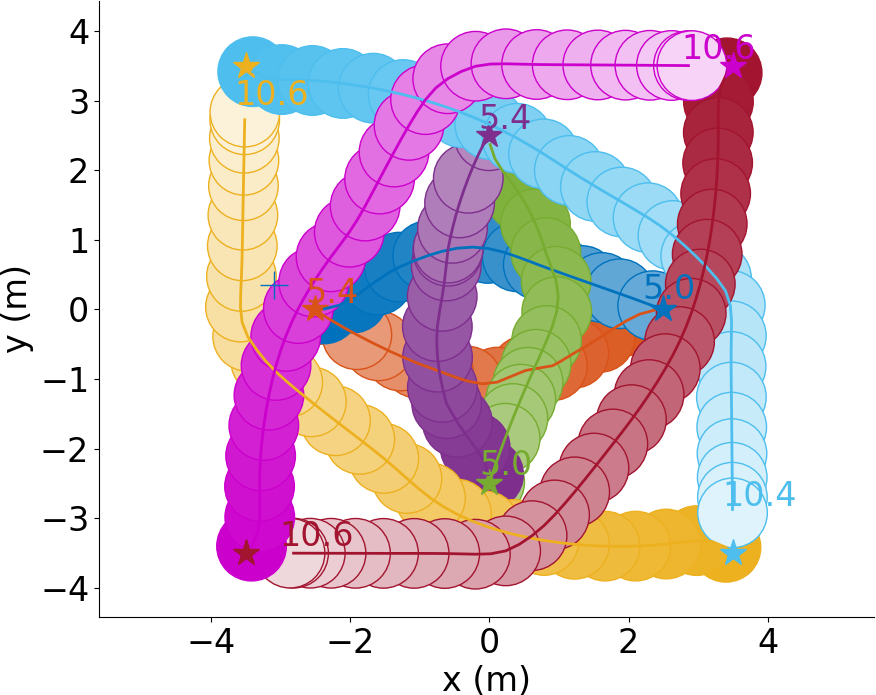}
    	\caption{GO-MPC}
   	\label{fig:homo_agents_mpcrl}
    \end{subfigure}%
    \caption{8 agents swapping positions. To simulate a multi-robot environment, all agents follow the same policy.}
    \label{fig:homo_agents}
\end{figure}

%\begin{figure*}[!t]
%  \centering
%  \begin{minipage}{\textwidth}
%    \includegraphics[height=5cm,width=17.8cm,trim={0cm 0cm 0cm 0cm},clip]{images/mixed_scenarios.png} % scene5_step_11
%    \caption*{\textit{a)} This sub-figure illustrates three sample results obtained for the asymmetric swapping scenario with mixed-agents, cooperative and non-cooperative. Green depicts the cooperative agents, red the non-cooperative and blue the ego-agent.}
%    \label{fig:mix_agents1} 
%  \end{minipage}
%   \vspace{1mm}

%    \begin{minipage}{\textwidth}
%      \includegraphics[height=5cm,width=17.8cm,trim={0cm 0cm 0cm 0cm},clip]{images/homogeneous_scenarios.png} %scene12_step_22 scene12_step_20 is a good one 13_8 too
%      \caption*{\textit{b)}  Three examples of the asymmetric swapping scenario where all the agents following our navigation policy. Each agent is assigned a different color. The goal position is depicted with a star symbol of the agent's color.}
%      \label{fig:homo_agents1}
%    \end{minipage}
%    \hspace{10mm}
%  \caption{ Split in two images. add also the analysis path among cooperative vs safe path among non cooperative}Simulation results for the asymmetric swapping scenario with mixed and homogeneous agents and,  for $n \in \{6,8,10\}$ agents.
%  \meXX{Let's replace 1 of the bottom right 2 examples to be one where the MPC-RL agents go more straight to their goals.}
%  }
%  \label{fig:qualitative_results}
%\end{figure*}   

We present results for mixed settings in \cref{fig:mix_agents} and homogeneous settings in \cref{fig:homo_agents} with $n \in \{6,8,10\}$ agents. In mixed settings, the robot follows our proposed policy while the other agents either follow an RVO \cite{van2008reciprocal} or a non-cooperative policy (same distribution as in training). \cref{fig:mix_agents} demonstrates that our navigation policy behaves differently when dealing with only cooperative agents or both cooperative and non-cooperative. Whereas in \cref{fig:mix_agents_cooperative} the robot navigates through the crowd, \cref{fig:mix_agents_mix} shows that the robot takes a longer path to avoid the congestion.

In the homogeneous setting, all agents follow our proposed policy.
\cref{fig:homo_agents} shows that our method achieves faster time-to-goal than two DRL baselines. Note that this scenario was never introduced during the training phase, nor have the agents ever experienced other agents with the same policy before. Following the DRL policy (\cref{fig:homo_agents_rl}), all agents navigate straight to their goal positions leading to congestion in the center with reactive avoidance. The trajectories from the DRL-2 approach (\cref{fig:homo_agents_rl_d2}) are more conservative, due to the limited acceleration available. In contrast, the trajectories generated by our approach (\cref{fig:homo_agents_mpcrl}), present a balance between going straight to the goal and avoiding congestion in the center, allowing the agents to reach their goals faster and with smaller distance traveled.

\subsection{Performance Results}\label{sec:results_dynamic}

This section aggregates performance of the various methods across 200 random scenarios.
Performance is quantified by average time to reach the goal position, percentage of episodes that end in failures (either collision or timeout), and the average distance traveled.

%Furthermore, to evaluate the guidance policy, we compare our method, GO-MPC, with a variant without collision avoidance constraints on, the GO-MPC-Off.

The numerical results are summarized in \cref{tab:global_performance}. Our method outperforms each baseline for both mixed and homogeneous scenarios. To evaluate the statistical significance, we performed pairwise Mann–Whitney U-tests between GO-MPC and each baseline (95$\%$ confidence). GO-MPC shows statistically significant performance improvements over the DRL-2 baseline in terms of travel time and distance, and the DRL baseline in term of travel time for six agents and travel distance for ten agents. %All the other results were not statistically significant
.
%Although MPC has the minimum average time to goal and traveled distance for the mixed agent's scenario, its collision avoidance performance is poor. Moreover, for the homogeneous scenario, MPC is unperformed by all methods in all metrics.
For homogeneous scenarios, GO-MPC is more conservative than DRL and MPC baselines resulting in a larger average traveled distance.
Nevertheless, GO-MPC is reaches the goals faster than each baseline and is less conservative than DRL-2, as measured by a significantly lower average distance traveled.
%For the homogeneous scenario, our approach is more conservative than the DRL~\cite{everett2019collision} and MPC baselines resulting in a larger average traveled distance.
%Nevertheless, our method is more time-efficient than all baselines and less conservative than the DRL-2~\cite{everett2019collision} method achieving a lower average travel distance.% in terms of traveled distance, and achieves the minimum average time to goal 

Finally, considering higher-order dynamics when training DRL agents (DRL-2) improves the collision avoidance performance. However, it also increases the average time to goal and traveled distance, meaning a more conservative policy that still under-performs GO-MPC in each metric.

%Finally, the results obtained for the GO-MPC-Off baseline show that for more than six agents, the learned guidance policy improves the collision avoidance performance in comparison with the pure MPC approach. Nevertheless, combining the MPC's collision avoidance performance with the guidance provided by the subgoal policy significantly improves the performance of our method.

\begin{table*}[!t]
 \caption{Statistics for 200 runs of proposed method (GO-MPC) compared to baselines (MPC, DRL \cite{everett2019collision} and DRL-2, an extension of \cite{everett2019collision}): time to goal and traveled distance for the successful episodes, and number of episodes resulting in collision for $n \in \{6,8,10\}$ agents. For the mixed setting, 80\% of agents are cooperative, and 20\% are non-cooperative.}
        \centering
\begin{tabular}{|p{1cm}|p{1.8cm}|p{1.8cm}|p{1.8cm}|p{1.8cm}|p{1.8cm}|p{1.8cm}|p{1.8cm}|p{1.8cm}|p{1.8cm}|p{1.8cm}|p{1.8cm}|p{1.8cm}|}
\hline
\multicolumn{2}{|c|}{} & \multicolumn{3}{c|}{Time to Goal (mean $\pm$ std) [s]} & \multicolumn{3}{|c|}{\% failures (\% collisions / \% deadlocks) } & \multicolumn{3}{c|}{Traveled Distance (mean $\pm$ std) [m]} \\ \hline
\multicolumn{2}{|c|}{\# agents}                    & \multicolumn{1}{c|}{6}                & \multicolumn{1}{c|}{8}       & \multicolumn{1}{c|}{10}                        & \multicolumn{1}{c|}{6}                &\multicolumn{1}{c|}{8}     & \multicolumn{1}{c|}{10}                       & \multicolumn{1}{c|}{6}                & \multicolumn{1}{c|}{8}       &\multicolumn{1}{c|}{10}    \\ \hline \hline %\multicolumn{11}{l}{\bf Mixed Agents - First Order Dynamics} \\  \hline \hline

 \multicolumn{11}{l}{\bf Mixed Agents } \\  \hline \hline
\multicolumn{2}{|c|}{MPC}                            & \multicolumn{1}{c|}{11.2 $\pm$ 2.2}      &  \multicolumn{1}{c|}{11.3 $\pm$ 2.4}          &     \multicolumn{1}{c|}{ 11.0 $\pm$ 2.2}          & \multicolumn{1}{c|}{13 (0 / 0)}       & \multicolumn{1}{c|}{ 22 (0 / 0)}     &   \multicolumn{1}{c|}{22 (22 / 0) }            &  \multicolumn{1}{c|}{ 12.24 $\pm$ 2.3}                &   \multicolumn{1}{c|}{12.40 $\pm$ 2.5}       & \multicolumn{1}{c|}{12.13 $\pm$ 2.3}   \\ \hline
\multicolumn{2}{|c|}{DRL~\cite{everett2019collision}}                      &  \multicolumn{1}{c|}{13.7 $\pm$ 3.0}     &  \multicolumn{1}{c|}{13.7 $\pm$ 3.1}       &    \multicolumn{1}{c|}{14.4 $\pm$ 3.3}         & \multicolumn{1}{c|}{ 17 (17 / 0)}    &  \multicolumn{1}{c|}{23 (23 / 0)}        &  \multicolumn{1}{c|}{29 (29 / 0)  }        &  \multicolumn{1}{c|}{13.75 $\pm$ 3.3}   &   \multicolumn{1}{c|}{13.80 $\pm$ 4.0}       & \multicolumn{1}{c|}{14.40 $\pm$ 3.3}    \\ \hline
\multicolumn{2}{|c|}{DRL-2~\cite{everett2019collision}+}                      &  \multicolumn{1}{c|}{15.3 $\pm$ 2.3}    & \multicolumn{1}{c|}{16.1 $\pm$ 2.2}        &    \multicolumn{1}{c|}{16.7 $\pm$ 2.2}          & \multicolumn{1}{c|}{ 6 (6 / 0)}     & \multicolumn{1}{c|}{10 (10 / 0)}         &  \multicolumn{1}{c|}{13 (13 / 0)}           & \multicolumn{1}{c|}{14.86 $\pm$ 2.3}    &  \multicolumn{1}{c|}{16.05 $\pm$ 2.2}        & \multicolumn{1}{c|}{16.66 $\pm$ 2.2}     \\ \hline
\multicolumn{2}{|c|}{GO-MPC}                &   \multicolumn{1}{c|}{12.7 $\pm$ 2.7}              &   \multicolumn{1}{c|}{12.9  $\pm$ 2.8 }            &  \multicolumn{1}{c|}{13.3  $\pm$ 2.8}       & \multicolumn{1}{c|}{\textbf{ 0 (0 / 0) }}    &  \multicolumn{1}{c|}{\textbf{ 0 (0 / 0)   } }       & \multicolumn{1}{c|}{\textbf{ 0 (0 / 0)   }}      & \multicolumn{1}{c|}{13.65  $\pm$ 2.7}                &\multicolumn{1}{c|}{13.77  $\pm$ 2.8}    &      \multicolumn{1}{c|}{14.29  $\pm$ 2.8}         \\ \hline \hline \multicolumn{11}{l}{\bf Homogeneous} \\  \hline \hline
\multicolumn{2}{|c|}{MPC}                &   \multicolumn{1}{c|}{17.37  $\pm$ 2.9}              &    \multicolumn{1}{c|}{16.38  $\pm$ 1.5}          &  \multicolumn{1}{c|}{16.64  $\pm$ 1.7}          &  \multicolumn{1}{c|}{ 30 (29 / 1)}     &  \multicolumn{1}{c|}{36 (25 / 11)}          &   \multicolumn{1}{c|}{ 35 (28 / 7)}              & \multicolumn{1}{c|}{ 11.34  $\pm$ 2.1}      & \multicolumn{1}{c|}{10.86  $\pm$ 2.3}    & \multicolumn{1}{c|}{10.62  $\pm$ 2.8}        \\ \hline
\multicolumn{2}{|c|}{DRL~\cite{everett2019collision}}                &  \multicolumn{1}{c|}{ 14.18  $\pm$ 2.4}           &     \multicolumn{1}{c|}{14.40  $\pm$ 2.7}                   &    \multicolumn{1}{c|}{14.64  $\pm$ 3.3}                &   \multicolumn{1}{c|}{16 (14 / 2) }               &   \multicolumn{1}{c|}{20 (18 / 2)}                  & \multicolumn{1}{c|}{20 (20 / 0) }      &   \multicolumn{1}{c|}{12.81  $\pm$ 2.3}               &   \multicolumn{1}{c|}{12.23  $\pm$ 2.3}               &    \multicolumn{1}{c|}{12.23  $\pm$ 3.2}          \\ \hline
\multicolumn{2}{|c|}{DRL-2~\cite{everett2019collision}+}                &  \multicolumn{1}{c|}{ 15.96  $\pm$ 3.1}           &     \multicolumn{1}{c|}{17.47  $\pm$ 4.2}                   &   \multicolumn{1}{c|}{15.96  $\pm$ 4.5}                   &  \multicolumn{1}{c|}{17 (11 / 6)}                 &    \multicolumn{1}{c|}{ 29 (21 / 8)}                & \multicolumn{1}{c|}{28 (24 / 4)}       &   \multicolumn{1}{c|}{15.17  $\pm$ 3.0}              &   \multicolumn{1}{c|}{15.85  $\pm$ 4.2}               &     \multicolumn{1}{c|}{15.40  $\pm$ 4.5}       \\ \hline
%\multicolumn{2}{|c|}{GO-MPC-Off}                &  \multicolumn{1}{c|}{14.08}                   &    \multicolumn{1}{c|}{15.54}                    &    \multicolumn{1}{c|}{15.89}                 &   \multicolumn{1}{c|}{19 (17 / 2) }             &    \multicolumn{1}{c|}{29 (29 / 0)}            &   \multicolumn{1}{c|}{26 (26 / 0)}          &   \multicolumn{1}{c|}{14.25}           &  \multicolumn{1}{c|}{15.45}          & \multicolumn{1}{c|}{15.94}      \\

\multicolumn{2}{|c|}{GO-MPC}                &   \multicolumn{1}{c|}{\textbf{13.77  $\pm$ 2.9}}         &    \multicolumn{1}{c|}{\textbf{14.30  $\pm$ 3.3}}         &    \multicolumn{1}{c|}{\textbf{14.63  $\pm$ 2.9}}         &  \multicolumn{1}{c|}{\textbf{  0 (0 / 0)  }} &  \multicolumn{1}{c|}{\textbf{  0 (0 / 0)  } }    &   \multicolumn{1}{c|}{ \textbf{ 2 (1 / 1)    }}   &    \multicolumn{1}{c|}{14.67  $\pm$ 2.9}         &  \multicolumn{1}{c|}{15.09  $\pm$ 3.3}         & \multicolumn{1}{c|}{15.12  $\pm$ 2.9}      \\
\hline
\end{tabular}
 \label{tab:global_performance}
\end{table*}

%\begin{figure}
%    \centering
%    \includegraphics[width=\columnwidth]{images/collision_avoidance_comparison.png}
%    \caption{Collision avoidance performance of our proposed method vs two state-of-the-art baselines: GA3C \cite{everett2018motion} and CADRL \cite{chen2017socially}}
%    \label{fig:comparison_baselines}
%\end{figure}

%\FloatBarrier
\section{CONCLUSIONS \& FUTURE WORK}\label{sec:conclusions}
This paper introduced a subgoal planning policy for guiding a local optimization planner. We employed DRL methods to learn a subgoal policy accounting for the interaction effects among the agents. Then, we used an MPC to compute locally optimal motion plans respecting the robot dynamics and collision avoidance constraints. Learning a subgoal policy improved the collision avoidance performance among cooperative and non-cooperative agents as well as in multi-robot environments. Moreover, our approach can reduce travel time and distance in cluttered environments. Future work could account for environment constraints.% and hardware experiments.

%%%%%%%%%%%%%%%%%%%%%%%%%%%%%%%%%%%%%%%%%%%%%%%%%%%%%%%%%%%%%%%%%%%%%%%%%%%%%%%%

%%%%%%%%%%%%%%%%%%%%%%%%%%%%%%%%%%%%%%%%%%%%%%%%%%%%%%%%%%%%%%%%%%%%%%%%%%%%%%%%

%%%%%%%%%%%%%%%%%%%%%%%%%%%%%%%%%%%%%%%%%%%%%%%%%%%%%%%%%%%%%%%%%%%%%%%%%%%%%%%%

%%%%%%%%%%%%%%%%%%%%%%%%%%%%%%%%%%%%%%%%%%%%%%%%%%%%%%%%%%%%%%%%%%%%%%%%%%%%%%%%

%\newpage
%\balance
%%%%%%%%%%%%%%%%%%%%%%%%%%%%%%%%%%%%%%%%%%%%%%%%%%%%%%%%%%%%%%%%%%%%%%%%%%
\bibliographystyle{IEEEtran}
\bibliography{MyBib}

\end{document}